\newcommand{\final}{1}
\newcommand{\arxiv}{1}
\documentclass[sigconf,nonacm]{acmart} 

\acmSubmissionID{249}

\usepackage[ruled]{algorithm2e} 

\SetAlFnt{\small}
\SetAlCapFnt{\small}
\SetAlCapNameFnt{\small}
\SetAlCapHSkip{0pt}

\usepackage{amsthm}

\usepackage{amssymb}
\usepackage{textcomp}
\usepackage{wrapfig}
\usepackage{subfig}
\usepackage{color}
\usepackage{xspace}
\usepackage{subfig}
\usepackage{enumitem}
\usepackage{hyperref}
\usepackage{multirow}
\usepackage{soul}
\usepackage{ifthen}
\usepackage{microtype}
\usepackage{bbding}
\usepackage{makecell}
\usepackage{bbold}

\newcommand{\revision}[1]{{\color{orange} #1}}

\ifthenelse{\equal{\final}{1}}
{
\renewcommand{\revision}[1]{#1}
}
{}

\newcommand{\sysname}{I2V-Adapter\xspace}

\newcommand{\gaussiannoise}{\boldsymbol{\epsilon}}
\newcommand{\neuralnetwork}{\boldsymbol{\epsilon}_{\theta}}

\copyrightyear{2024}
\acmYear{2024}
\setcopyright{acmlicensed}\acmConference[SIGGRAPH Conference Papers '24]{Special Interest Group on Computer Graphics and Interactive Techniques Conference Conference Papers '24}{July 27-August 1, 2024}{Denver, CO, USA}
\acmBooktitle{Special Interest Group on Computer Graphics and Interactive Techniques Conference Conference Papers '24 (SIGGRAPH Conference Papers '24), July 27-August 1, 2024, Denver, CO, USA}
\acmDOI{10.1145/3641519.3657407}
\acmISBN{979-8-4007-0525-0/24/07}

\begin{document}

\title{I2V-Adapter: A General Image-to-Video Adapter for \\ Diffusion Models}

\ifthenelse{\equal{\arxiv}{1}}
{
\author{Xun Guo$^{1*\dagger}$\hspace{0.1in} Mingwu Zheng$^{2*}$\hspace{0.1in} Liang Hou$^{2}$\hspace{0.1in} Yuan Gao$^2$\hspace{0.1in} Yufan Deng$^2$\hspace{0.1in} Pengfei Wan$^{2}$ \\ Di Zhang$^{2}$ \hspace{0.1in} Yufan Liu$^{3}$\hspace{0.1in} Weiming Hu$^{3}$\hspace{0.1in} Zhengjun Zha$^{1}$\hspace{0.1in} Haibin Huang$^{2\ddagger}$\hspace{0.1in} Chongyang Ma$^{2}$ \vspace{4pt}\\
\institution{$^1$University of Science and Technology of China \hspace{0.2in} $^2$Kuaishou Technology\\
$^3$Institute of Automation, Chinese Academy of Sciences}}
}
{
\author{Xun Guo}
\affiliation{
\institution{University of Science and Technology of China}
\country{China}
}
\email{xunguo@mail.ustc.edu.cn}

\author{Mingwu Zheng}
\affiliation{
\institution{Kuaishou Technology}
\country{China}
}
\email{zhengmingwu12@gmail.com}

\author{Liang Hou}
\affiliation{
\institution{Kuaishou Technology}
\country{China}
}
\email{lianghou96@gmail.com}

\author{Yuan Gao}
\affiliation{
\institution{Kuaishou Technology}
\country{China}
}
\email{ggarysm@gmail.com}

\author{Yufan Deng}
\affiliation{
\institution{Kuaishou Technology}
\country{China}
}
\email{dyflional10@gmail.com}

\author{Pengfei Wan}
\affiliation{
\institution{Kuaishou Technology}
\country{China}
}
\email{wanpengfei@kuaishou.com}

\author{Di Zhang}
\affiliation{
\institution{Kuaishou Technology}
\country{China}
}
\email{zhangdi08@kuaishou.com}

\author{Yufan Liu}
\affiliation{
\institution{Institute of Automation, Chinese Academy of Sciences}
\country{China}
}
\email{yufan.liu@ia.ac.cn}

\author{Weiming Hu}
\affiliation{
\institution{Institute of Automation, Chinese Academy of Sciences}
\country{China}
}
\email{wmhu@nlpr.ia.ac.cn}

\author{Zhengjun Zha}
\affiliation{
\institution{University of Science and Technology of China}
\country{China}
}
\email{zhazj@ustc.edu.cn}

\author{Haibin Huang}
\authornote{Corresponding author.}
\affiliation{
\institution{Kuaishou Technology}
\country{China}
}
\email{jackiehuanghaibin@gmail.com}

\author{Chongyang Ma}
\affiliation{
\institution{Kuaishou Technology}
\country{China}
}
\email{chongyangm@gmail.com}
}

\renewcommand{\shorttitle}{I2V-Adapter: A General Image-to-Video Adapter for Diffusion Models}
\renewcommand{\shortauthors}{Guo et al.}
\renewcommand{\authors}{Xun Guo, Mingwu Zheng, Liang Hou, Yuan Gao, Yufan Deng, Pengfei Wan, Di Zhang, Yufan Liu, Weiming Hu, Zhengjun Zha, Haibin Huang, and Chongyang Ma}

\begin{abstract}
\ifthenelse{\equal{\arxiv}{1}}
{
\renewcommand{\thefootnote}{}
\footnotetext{$^{*}$ Joint first authors.}
\footnotetext{$^{\dagger}$ Work done as an intern at Kuaishou Technology.}
\footnotetext{$^{\ddagger}$ Corresponding author: jackiehuanghaibin@gmail.com}
}
{}
Text-guided image-to-video (I2V) generation aims to generate a coherent video that preserves the identity of the input image and semantically aligns with the input prompt. Existing methods typically augment pretrained text-to-video (T2V) models by either concatenating the image with noised video frames channel-wise before being fed into the model or injecting the image embedding produced by pretrained image encoders in cross-attention modules. However, the former approach often necessitates altering the fundamental weights of pretrained T2V models, thus restricting the model's compatibility within the open-source communities and disrupting the model's prior knowledge. Meanwhile, the latter typically fails to preserve the identity of the input image. We present I2V-Adapter to overcome such limitations. I2V-Adapter adeptly propagates the unnoised input image to subsequent noised frames through a cross-frame attention mechanism, maintaining the identity of the input image without any changes to the pretrained T2V model. Notably, I2V-Adapter only introduces a few trainable parameters, significantly alleviating the training cost and also ensures compatibility with existing community-driven personalized models and control tools. Moreover, we propose a novel Frame Similarity Prior to balance the motion amplitude and the stability of generated videos through two adjustable control coefficients. Our experimental results demonstrate that I2V-Adapter is capable of producing high-quality videos. This performance, coupled with its agility and adaptability, represents a substantial advancement in the field of I2V, particularly for personalized and controllable applications.

\end{abstract}

%
%



\newcommand{\todo}[1]{{\it\color{red} TODO: #1}}
\newcommand{\eg}{{\it e.g.}}
\newcommand{\ie}{{\it i.e.}}

\maketitle
\begin{figure*}
	\includegraphics[width=\textwidth]{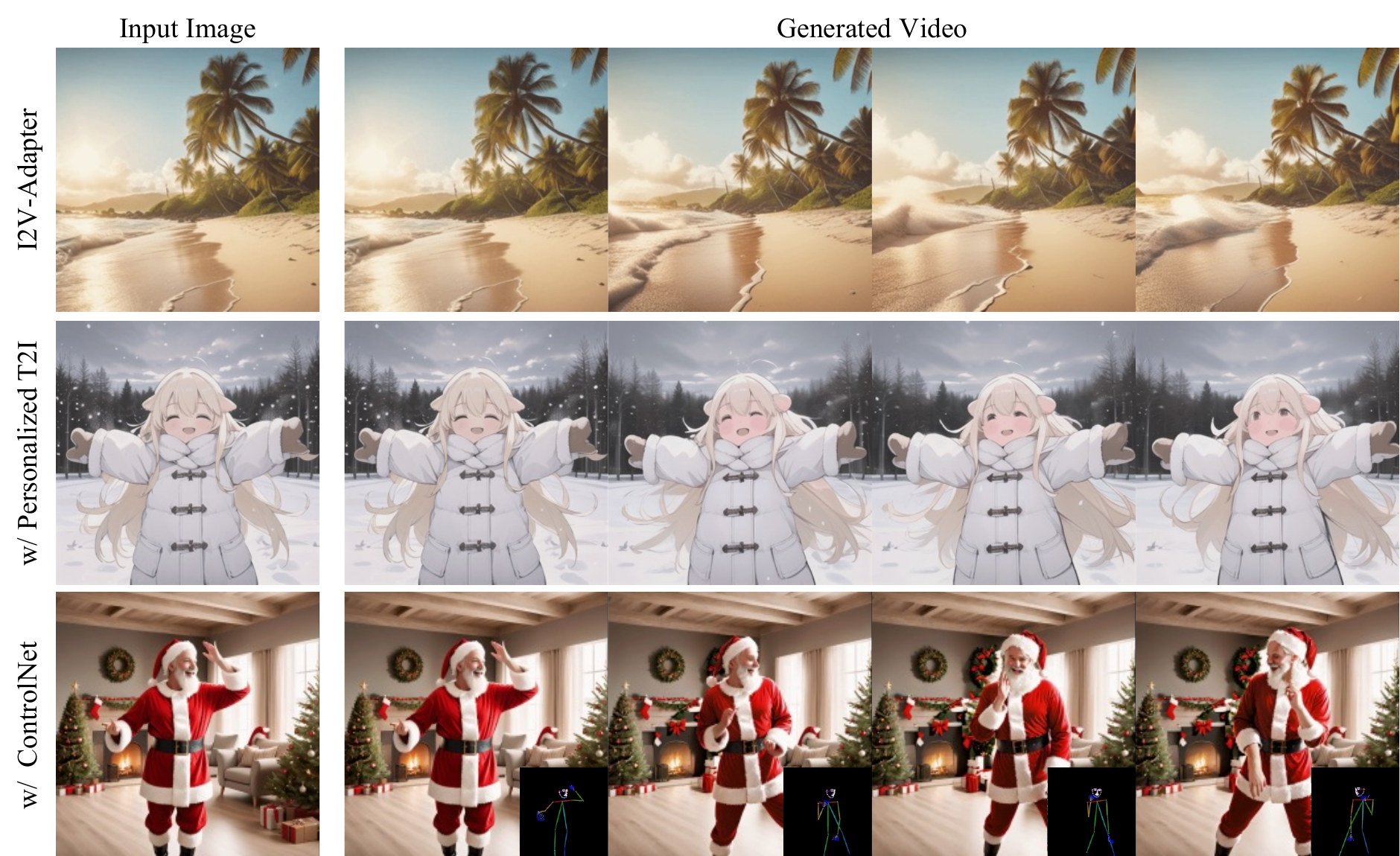}
	\caption{
	\label{fig:teaser}
        Our lightweight I2V-Adapter achieves high-quality and versatile image-to-video generation, \textit{without tuning any pretrained spatial or temporal modules}. \textbf{{First row:}} I2V results of I2V-Adapter. \textbf{{Second row:}} results of I2V-Adapter combined with a personalized T2I model (\href{https://civitai.com/models/28169/cuteyukimixadorable-style?modelVersionId=265102}{\textit{CuteYukiMix}}). \textbf{{Third row:}} results of I2V-Adapter combined with ControlNet~\cite{controlnet}.
    }
	\Description[TeaserFigure]{TeaserFigure}
\end{figure*}
\section{Introduction}

As an essential medium, video has risen to prominence on the Internet, offering users a rich and dynamic visual experience. The technology behind video generation has therefore garnered widespread attention due to its capacity to create innovative video content and its immense potential in the field of generative artificial intelligence. 
Recently, with the remarkable success of text-to-image (T2I) diffusion models such as Stable Diffusion~\cite{stablediffusion} and SDXL~\cite{SDXL}, there have been significant advancements in text-to-video (T2V) diffusion models~\cite{ho2022imagenvideo, emuvideo, chen2023videocrafter1, guo2023animatediff, svd, Mullan_Hotshot-XL_2023, wang2023modelscope, wang2023lavie}. These models integrate temporal modules, such as temporal convolutions or temporal self-attention, onto T2I diffusion models to capture the relations between video frames. 
Typical approaches like AnimateDiff~\cite{guo2023animatediff} decouples spatial modeling and temporal modeling to learn a motion prior from video data in a plug-and-play manner.
By solely training the newly introduced temporal modules while maintaining the pretrained weights of T2I models, these approaches have achieved significant improvements in the field of T2V.

While T2V models have shown significant progress, crafting effective text prompt to generate the desired content remains a challenge, often necessitating complex prompt engineering~\cite{witteveen2022investigating}. 
This intricacy highlights the advantages of an alternative methodology, \ie, image-to-video (I2V) generation, which takes an image together with a text prompt as the input.
This kind of method is often more intuitive and convenient, since the input image provides a direct and tangible visual reference for the video generation.
However, translating specific images into coherent frame sequences in high fidelity necessitate not only generative capabilities but also fine-grained understanding of input images. 

To tackle this unique challenge, many existing I2V models consider how to effectively inject the input image into the diffusion process during training, building upon the pretrained T2V models. One common approach involves channel-wise concatenation of the input image with the original noised inputs of the diffusion model's U-Net~\cite{svd, emuvideo, chen2023livephoto, zeng2023makepixeldance}. Such a modification inevitably necessitates alterations to the original model's structure and weights as the input space undergoes significant changes. Typically, this approach requires intensive training on a voluminous video-text dataset to mitigate this discrepancy. However, such an approach exacerbates the complexity and instability of the training process which could result in catastrophic forgetting and significantly hampers the performance and generalizability.
Moreover, this modification limits their reusability with personalized T2I models~\cite{DreamBooth, lora} as well as their compatibility with control tools like ControlNet~\cite{controlnet}.
Notably, these models and tools, developed by a vast community of AI artists and enthusiasts, have been instrumental in numerous applications. Another common approach explored is the additional cross attention mechanism that relates the outputs of a pretrained image encoder (\eg, CLIP~\cite{clip} and DINO~\cite{oquab2023dinov2}) with the diffusion model~\cite{chen2023livephoto, 2023i2vgenxl, chen2023videocrafter1}. This strategy leverages the ability of cross attention mechanism to capture high-level dependencies and correlations. However, the outputs of these encoders often fall short in preserving the identity and fine details of the input image, leading to results that are less than satisfactory.

In this work, we present \sysname, a lightweight and plug-and-play adapter designed to overcome aforementioned limitations for image-to-video generation. \sysname capitalizes on the prior knowledge and capabilities of the pretrained T2V model by feeding the unnoised input image \textit{in parallel with} the subsequent noised frames to the model. Our approach propagates the input image's identity information to the subsequent frames through a cross-frame attention mechanism without altering any parameters of the spatial or motion modules (Stable Diffusion and AnimateDiff in our implementation). \sysname is integrated with the pretrained model via a trainable copy of query projector and a trainable zero-initialized output projector, ensuring that the model's initialization remains unaffected by the newly incorporated modules. Notably, \sysname introduces only a few trainable parameters, significantly alleviating the training cost while ensuring the compatibility with the open-source communities. Moreover, we propose a novel Frame Similarity Prior to balance the motion magnitude and stability in generated videos.



To summarize, our contributions are as follows:

\begin{itemize}[leftmargin=*]
\item We present \sysname, a novel lightweight adapter that achieves general image-to-video generation without training any pre-existing spatial or motion modules.

\item We propose the Frame Similarity Prior, a novel concept that leverages the inherent similarity between video frames for image-to-video generation. This approach is designed to balance the stability with motion magnitude in generated videos, thereby enhancing the model's controllability and the diversity of generated videos.

\item Our experiments demonstrate that our model can achieve superior results in image-to-video generation. Notably, our model requires significantly fewer trainable parameters (down to a minimum of 24M), as low as \textit{1\%} of mainstream models.

\item We show that our \sysname exhibits versatility and compatibility, supporting a variety of personalized T2I models and controllable tools, thus enabling creative and custom applications.

\end{itemize}

\section{Related Work}
\label{sec:related}

Visual content generation, notably influenced by the advancements in diffusion models~\cite{ho2020denoising, song2020score, song2020denoising}, is emerging as a popular research area.
This section provides an overview of closely related work from three perspectives: text-to-video (T2V) generation, image-to-video (I2V) generation, and adapter techniques.

\paragraph{Text-to-video generation.}
T2V generation extends beyond the capabilities of T2I by producing a sequence of continuous frames, where diffusion models also play a pivotal role. Unlike T2I, T2V demands the synthesis of temporally coherent and visually consistent video content from textual descriptions, posing unique challenges in maintaining continuity and realism over time.

In the realm of T2V, early methods such as LVDM~\cite{he2022lvdm} and ModelScope~\cite{wang2023modelscope} have pioneered the integration of temporal modules with the 2D U-Net~\cite{ronneberger2015unet} architecture. These approaches often involve redesign and comprehensive training of volumetric U-Net models, incorporating temporal aspects to cater to the dynamic nature of video content. This methodology signifies a critical step in handling the temporal complexity inherent in video generation. 
Another notable category of methods includes AnimateDiff~\cite{guo2023animatediff} and Hotshot-XL~\cite{Mullan_Hotshot-XL_2023}. These models incorporate trainable temporal layers while retaining the foundational structure of existing T2I frameworks. This strategy leverages the established efficacy of T2I models, extending their capabilities to handle the additional dimension of time, thereby facilitating the generation of seamless video sequences.
Additionally, approaches like Imagen Video~\cite{ho2022imagenvideo} and LaVie~\cite{wang2023lavie} adopt a cascaded structure with temporal and spatial super-resolution modules. This kind of architecture addresses both the spatial and temporal intricacies of video content, ensuring higher resolution and temporally coherent video generation.

To achieve multimodal compatibility and enhanced controllability within a unified framework, VideoComposer~\cite{wang2023videocomposer} emerges as a significant contribution. This model integrates various modalities, enabling more controllable video generation, a crucial factor in applications demanding high fidelity and specific contextual alignment.
\revision{SparseCtrl~\cite{guo2023sparsectrl} employs an additional encoder on top of the base T2V model which injects temporally sparse conditions (\eg, sketch, depth, RGB image) into the diffusion process, enabling the base model to be used for various applications, including sketch-to-video, depth-guided generation, and image animation.}

\paragraph{Image-to-video generation.}
I2V generation is characterized by its requirement to maintain the identity of the input image in the generated video. This poses a challenge in balancing the preservation of the image's identity with the dynamic nature of video.

Mask-based approaches like MCVD~\cite{voleti2022mcvd} and SEINE \cite{chen2023seine} utilize video prediction techniques to achieve I2V generation. These methods are pivotal in maintaining the continuity of the input content across the generated video frames, ensuring a seamless transition from the static to the dynamic.
Other innovations such as I2VGen-XL~\cite{2023i2vgenxl} and VideoCrafter1 \cite{chen2023videocrafter1} take a different approach by employing the CLIP~\cite{clip} image encoder. They construct image embeddings as conditions, enabling a more nuanced interpretation of the initial image and its subsequent transformation into video format.
Another set of techniques, including PixelDance~\cite{zeng2023makepixeldance}, Stable Video Diffusion~\cite{svd}, and VideoGen~\cite{li2023videogen}, operate in the VAE~\cite{kingma2013vae} latent space. They concatenate image data within this space, effectively incorporating the input image into the video generation framework. This method has shown promise in maintaining the fidelity of the input image while enabling dynamic video creation.
LAMP~\cite{wu2023lamp}, in particular, employs a first-frame-attention mechanism. This method focuses on transferring the information of the first frame to subsequent frames, ensuring that the essence of the initial image is reflected throughout the video sequence. While this method has shown promising results, its application is limited to fixed motion patterns in a few-shot setting. This limitation restricts its ability to generalize across a wide range of scenarios.
\revision{AnimateAnyone~\cite{hu2023animate} and MagicAnimate~\cite{xu2023magicanimate} both focus on creating human animations using pose sequences. AnimateAnyone utilizes a ReferenceNet for image encoding and a Spatial-Attention for identity consistency. It also utilizes a Pose Guider module to aid animation production. MagicAnimate employs a similar mechanism with an Appearance Encoder and ControlNet~\cite{controlnet} for image and pose encoding. The design of both systems resembles a dual-tower U-Net structure.}

In our work, we diverge from these singular approaches and introduce a method characterized by its plug-and-play nature. Our approach is designed to be compatible with various tools within the Stable Diffusion~\cite{stablediffusion} community, such as DreamBooth~\cite{DreamBooth}, LoRA~\cite{lora}, and ControlNet~\cite{controlnet}. This compatibility enhances the flexibility and applicability of our method. Furthermore, our approach is designed for ease of training, making it accessible for a broader range of applications and users.

\paragraph{Adapter.}
In the field of natural language processing, adapters \cite{houlsby2019adapter} have become a popular technique for efficient transfer and fine-tuning of pretrained large language models, such as BERT~\cite{bert} and GPT~\cite{radford2018gpt}. Adapters function by inserting small, task-specific modules between layers of pretrained models, thus enabling the learning of new task features without altering the original pretrained weights. The advantage of this approach is that it allows the model to retain foundational knowledge while rapidly adapting to specific tasks without the need for costly retraining of the entire network.

In the realm of diffusion models, a similar concept is being applied. ControlNet~\cite{controlnet} and T2I-Adapter~\cite{mou2023t2iadapter} are among the first to propose the use of adapters to integrate more control signals for diffusion models. IP-Adapter~\cite{ye2023ipadapter} specifically proposes a decoupled cross-attention mechanism for image conditioning, enabling the model to generate images that resemble the input image.

In our work, we introduce \sysname, a universal adapter that achieves efficient transfer from T2V to I2V with minimal overhead in terms of parameter number and works in a plug-and-play manner.

\section{Method}
\label{sec:method}

Given a reference image and a text prompt, our goal is to generate a video sequence starting with the provided image. This task is particularly challenging as it requires ensuring consistency with the first frame, compatibility with the prompt, and preserving the coherence of the video sequence throughout. 


\subsection{Preliminaries}
\label{sec:preliminaries}
\paragraph{Latent diffusion models.}
Diffusion models~\cite{ho2020denoising, song2020score, dhariwal2021diffusion, song2020denoising} are a class of generative models that involves a fixed forward diffusion process and a learned reverse denoising diffusion process.
Given a data $\mathbf{x}_0\sim p_\mathrm{data}(\mathbf{x})$, the forward process gradually adds Gaussian noise $\gaussiannoise\sim\mathcal{N}(\boldsymbol{0},\boldsymbol{1})$ to the data to obtain a noised data, as shown as follows:
\begin{equation}
    \mathbf{x}_t = \alpha_t\mathbf{x}_0 + \sigma_t\gaussiannoise,
\end{equation}
where $\alpha_t,\sigma_t\in[0,1]$ are two predefined functions of random time step $t\sim \mathcal{U}({1,\dots,T})$, satisfying the variance preserving principle, \ie, $\alpha_t^2 + \sigma_t^2=1$.

Diffusion models typically train a neural network $\neuralnetwork(\cdot)$ to learn the added noise, which is denoised during sampling to gradually generate new data.
For any condition $\mathbf{c}$ (\eg, text, image, and audio), the training objective function is defined as
\begin{equation}
    \min_{\theta} \mathbb{E}_{\mathbf{x}_0,\mathbf{c},t,\gaussiannoise}\left[\|\gaussiannoise-\neuralnetwork(\mathbf{x}_t,\mathbf{c},t)\|_2^2\right].
\end{equation}
Latent diffusion models (LDM) such as Stable Diffusion~\cite{stablediffusion} integrate the diffusion process within the low-dimensional latent space of a variational autoencoder (VAE)~\cite{kingma2013vae} instead of the high-dimensional pixel space for fast and efficient training and inference, which also yields comparable text-to-image generation performance.
The denoising network of diffusion models typically adopts a U-Net based architecture~\cite{ronneberger2015unet}, where each block contains residual convolution layers~\cite{he2016resnet}, self-attention blocks~\cite{vaswani2017attention}, and cross-attention blocks~\cite{vaswani2017attention}.



\paragraph{Self-attention in LDM}
The self-attention mechanism models the relationships between feature patches that generate the image. Self-attention has been demonstrated to control the layout, object shapes \cite{hertz2022p2p, tumanyan2023plug}, and \revision{fine-grained texture details~\cite{cao2023masactrl}} of generated images. Self-attention can be defined by:
\begin{equation}
\text{Attention}(\mathbf{Q}^i,\mathbf{K}^i,\mathbf{V}^i)=\text{softmax}\left(\frac{\mathbf{Q}^i (\mathbf{K}^i)^T}{\sqrt{d}}\right)\mathbf{V}^i,
\end{equation}
where $\mathbf{Q}^i = \mathbf{X}^{i} \mathbf{W}_Q$ represents the query from the input feature map $\mathbf{X}^{i}$ through the query weight matrix $\mathbf{W}_Q$, $\mathbf{K}^i = \mathbf{X}^{i} \mathbf{W}_K$ represents the key from the input feature map $\mathbf{X}^{i}$ through the key weight matrix $\mathbf{W}_K$, and $\mathbf{V}^i = \mathbf{X}^{i} \mathbf{W}_V$ represents the value from the input feature map $\mathbf{X}^{i}$ through the value weight matrix $\mathbf{W}_V$.

\paragraph{Cross-attention in LDM}
To incorporate textual conditional information into networks for text-to-image tasks, cross-attention has been introduced in LDM to model the dependency between image feature patches and textual tokens. In the cross-attention mechanism, the query $\mathbf{Q}^i = \mathbf{X}^{i} \mathbf{W}_Q^\text{cross}$ stems from the image feature $\mathbf{X}^i$, while the key $\mathbf{K}_\text{text} = \mathbf{X}_\text{text} \mathbf{W}_K^\text{cross}$ and the value $\mathbf{V}_\text{text} = \mathbf{X}_\text{text} \mathbf{W}_V^\text{cross}$ are derived from text embeddings $\mathbf{X}_\text{text}$ of textual tokens encoded via the CLIP text encoder~\cite{clip}.
This attention mechanism enables different textual tokens to govern the appearance of objects at different positions within the image. Cross-attention can be defined by the following equation:
\begin{equation}
\text{Attention}(\mathbf{Q}^i,\mathbf{K}_\text{text},\mathbf{V}_\text{text})=\text{softmax}\left(\frac{\mathbf{Q}^i \mathbf{K}_\text{text}^T}{\sqrt{d}}\right)\mathbf{V}_\text{text}.
\end{equation}

\subsection{\sysname}
\label{sec:i2v_adapter}

\paragraph{Temporal modeling with Stable Diffusion.}
The challenge in video generation lies in handling high-dimensional data to ensure coherence along the temporal dimension. Most existing video generation models incorporate motion modules on top of Stable Diffusion~\cite{stablediffusion} to model the temporal relationships in videos. In this paper, we are interested in AnimateDiff~\cite{guo2023animatediff}, a model initially developed for personalized T2V generation. It employs transformer blocks with positional encoding to capture the temporal relationships of the feature map along the dimension of frames. 
Due to this frame-wise attention mechanism, the motion module operates independently from the spatial modules, enabling it to generalize across multiple varieties of Stable Diffusion.



Inspired by this unique feature, we believe that the pretrained motion module can be seen as a universal motion representation, capable of being applied to more general video generation tasks, \eg, I2V generation, without any finetuning. Consequently, in this paper, we only modify \emph{spatial} features and keep the pretrained AnimateDiff temporal weights fixed.




\begin{figure}
    \centering
    \includegraphics[width=1.0\linewidth]{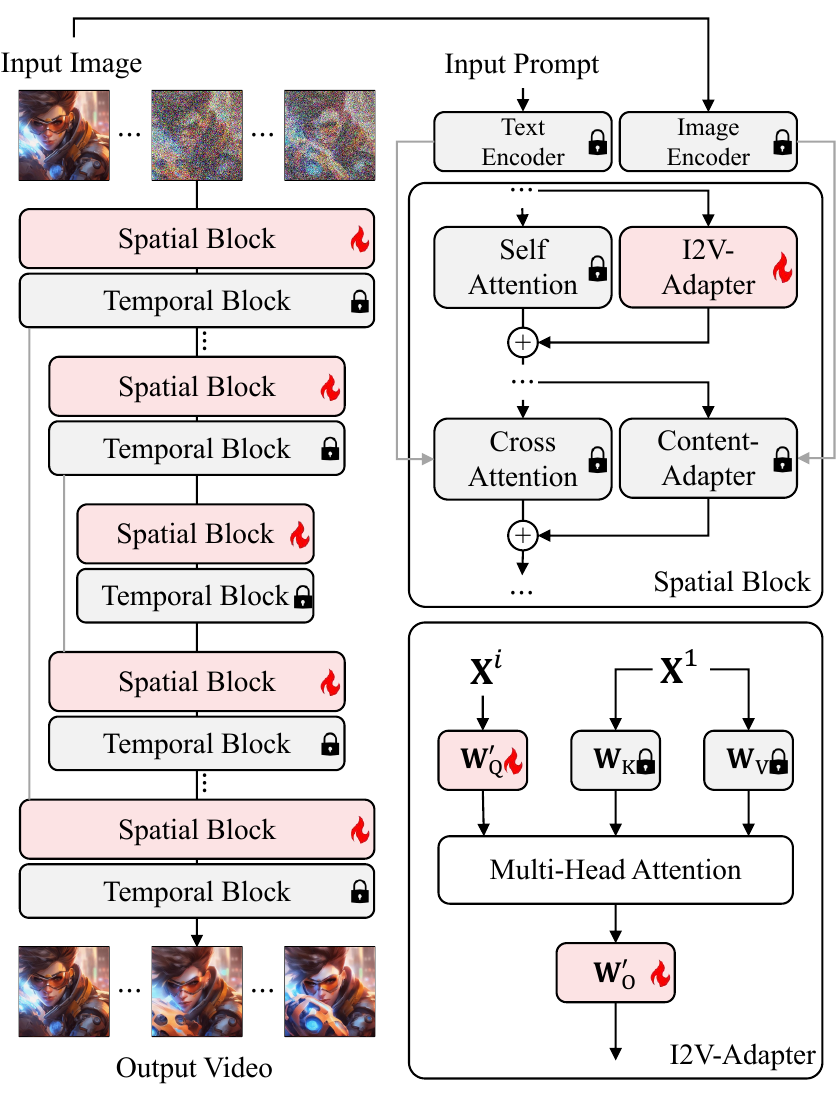}
    \caption{The overall architecture of our method.
    During both the training and inference procedures, we set the first frame as the input image and keep it noise-free.
    To capture fine-grained information, we add an additional branch alongside the self-attention block, \ie, \sysname. For each subsequent frame, the attention keys and values are shared with the first frame and remain unchanged. Only the query weights and zero-initialized output weights are trainable. This lightweight architecture allows us to infuse the input image into the diffusion process effectively.}
\label{fig:pipeline}
\end{figure}

\paragraph{Adapter for attention layers.}
Unlike text-to-video (T2V) generation, image-to-video (I2V) generation not only requires adherence to the high-level semantic information of a textual prompt but also needs to maintain the content information of the reference image, specifically its fine-grained, low-level features.
%
To this end, a common approach is to concatenate the reference image with the original noised inputs along the channel dimension before being fed into the U-Net~\cite{ronneberger2015unet}, thereby introducing reference image information at the input stage. Unfortunately, concatenation of additional input channels often necessitates retraining the entire U-Net network. This may disrupt the prior knowledge learned by the original Stable Diffusion model and also limits the flexibility to incorporate controllable or personalized models (\eg, ControlNet~\cite{controlnet} and DreamBooth~\cite{DreamBooth}) developed by the open-source community.

Unlike these approaches, we apply the first-frame-conditioned pipeline proposed by LAMP~\cite{wu2023lamp}, in which the first frame is always preserved as unnoised during both the training and inference stages, as detailed in Fig.~\ref{fig:pipeline}.
This pipeline keeps the space of the U-Net's input and output unchanged, thereby reducing the trainable parameters while protecting the pretrained model weights.


However, this condition is weak since it is challenging for the motion module to ensure interaction with the first frame. To enhance the condition, most existing methods propose using a pretrained image encoder, such as CLIP~\cite{clip} or DINO~\cite{oquab2023dinov2}, to extract features and then feed them into the cross-attention module of U-Net. We follow this paradigm and integrate a pretrained Content-Adapter~\cite{ye2023ipadapter} into the cross-attention module. However, despite the improvement, these image encoders usually only capture high-level semantic information and struggle to preserve lower-level features of the reference image. 

In this paper, we propose a novel and effective approach to fuse low-level information into the first-frame-conditioned pipeline. Our key observation is that the self-attention plays a critical role in controlling the layout, object shapes, as well as fine-grained spatial details during the generation process~\cite{hertz2022p2p, tumanyan2023plug}. Therefore, we add a cross-frame attention layer alongside the original self-attention layer to directly query the rich context information of the first frame encoded by the pretrained U-Net.

Specifically, for each frame $i \in \{1,...,l\}$, where $l$ indicates the video length, we add a new cross-frame attention layer to integrate the input image features into the original self-attention process. Given the key and value features of the first frame $\mathbf{K}^1$ and $\mathbf{V}^1$, the output of the new cross-frame attention is computed as follows:
\begin{equation}
\text{Attention}((\mathbf{Q}^i)',\mathbf{K}^1,\mathbf{V}^1)=\text{softmax}\left(\frac{(\mathbf{Q}^i)' (\mathbf{K}^1)^T}{\sqrt{d}}\right)\mathbf{V}^1,
\end{equation}
where $(\mathbf{Q}^i)' = \mathbf{X}^{i} \mathbf{W}'_Q$, and $\mathbf{W}'_Q$ represents a newly trainable query weight matrix. The parameters of $\mathbf{W}'_Q$ are initialized from the original $\mathbf{W}_Q$ in the corresponding layer. We then add the output to the original self-attention output, using the output projection matrices $\mathbf{W}_O$ and $\mathbf{W}'_O$:
{
\begin{equation}
\mathbf{X}^{i}_\text{out}\!=\!\text{Attention}(\mathbf{Q}^i,\mathbf{K}^i,\mathbf{V}^i) \mathbf{W}_O + \text{Attention}((\mathbf{Q}^i)',\mathbf{K}^1,\mathbf{V}^1) {\mathbf{W}'_O} .
\end{equation}
}

From a macroscopic perspective, the additional term on the right side borrows rich information from the first frame by querying its keys and using its values. Furthermore, the original self-attention module remains unchanged, ensuring that the T2I capability is not compromised. Besides, inspired by ControlNet~\cite{controlnet}, we zero-initialize the output projection layer $\mathbf{W}'_{O}$ to ensure that the model starts as if no modifications have been made. In total, our lightweight \sysname comprises only two trainable matrices: 
$\mathbf{W}'_Q$ and $\mathbf{W}'_O$, while leaving the pretrained weights of Stable Diffusion and the motion module completely unchanged, thereby ensuring that their original functionality remains optimal.

\subsection{Frame Similarity Prior}
\label{sec:frame_similariy_prior}

\begin{figure}
    \centering
    \includegraphics[width=1.0\linewidth]{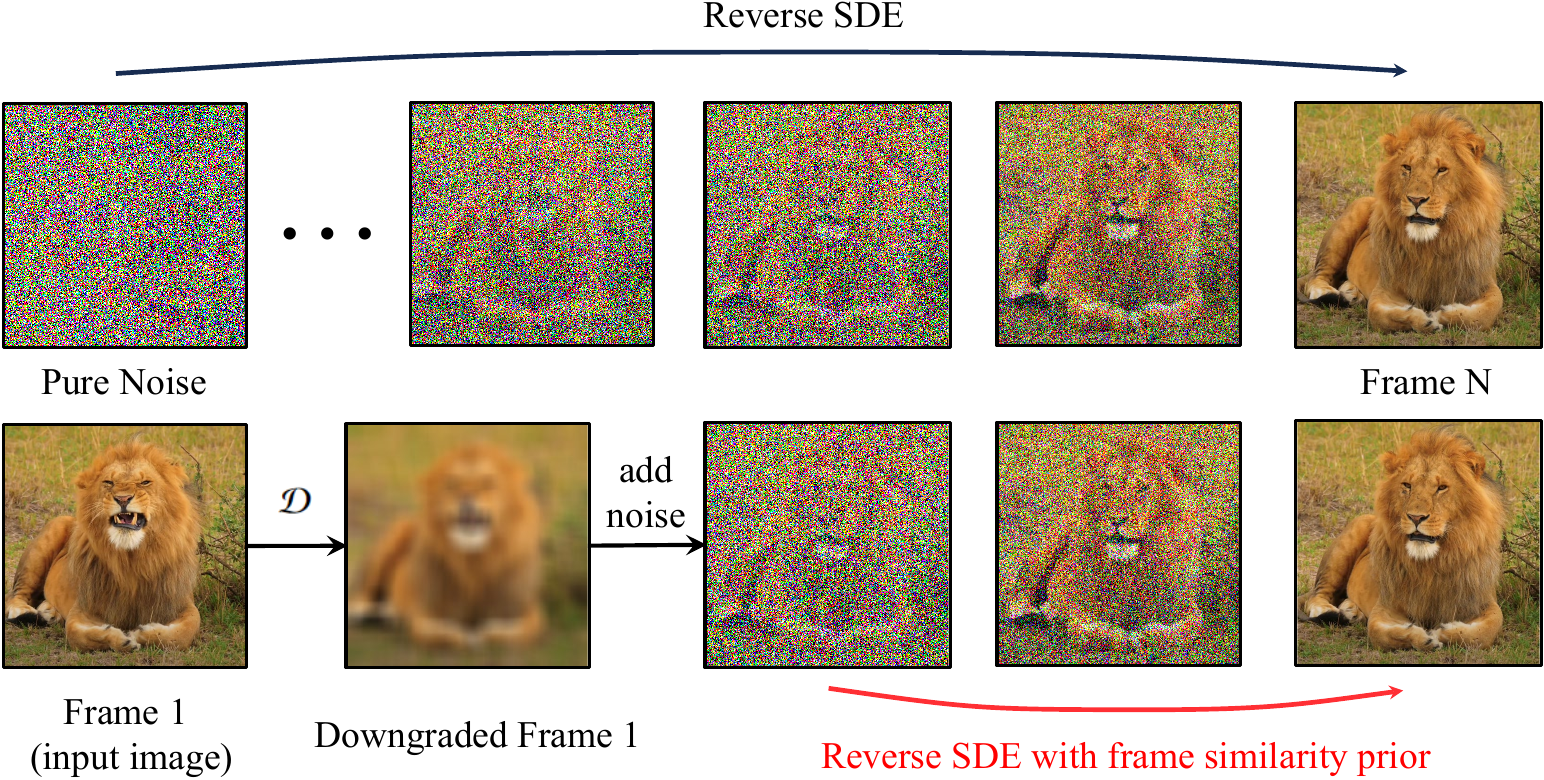}
    \caption{An illustration about Frame Similarity Prior. Instead of solving the reverse SDE from pure noise (the top row), we start the denoising process from corrupted input image (the bottom row). The frame similarity prior allows the model to generate videos with fewer steps and stabilizes the final result.}
\label{fig:sde}
\end{figure}

It is noteworthy that I2V generation is an ill-posed problem since the model is required to generate a full video from just a single reference image.
Despite the tremendous data prior we leverage from pretrained spatial and temporal modules, the model still occasionally tends to produce unstable and distorted results.

To tackle this problem, we propose an additional frame similarity prior. Our key assumption is that, on a relatively low Gaussian noise level, the marginal distributions of the noised first frame and the noised subsequent frames are adequately close.
At a high level, we assume that in most short video clips, all frames are structurally similar and become indistinguishable after being corrupted with a certain amount of Gaussian noise, see Fig.~\ref{fig:sde}. Specifically, inspired by SDEdit~\cite{sdedit}, to inject this prior into the diffusion process, instead of solving SDE~\cite{song2020score} from pure noise $\mathbf{x}_T$, we start from an intermediate time $t_0 \in (0, 1]$ and sample each frame $i$ as $\mathbf{x}^i_{t_0}\!\sim\!\mathcal{N}( \alpha_{t_0} \mathcal{D}({\mathbf{x}^1_0}); \sigma^2_{t_0}\mathbf{I})$.

To strike a balance between the realism and the strength of motion effects, we define the degradation operation $\mathcal{D}$ as a combination of masked Gaussian blur. Specifically, the operation is given by:
\begin{equation}
\mathcal{D}(\mathbf{x}) = \mathbf{M}_p \circ \mathbf{x} + (\mathbb{1}-\mathbf{M}_p) \circ \text{GaussianBlur}(\mathbf{x}), 
\end{equation}
where $\mathbf{M}_p$ is a randomly generated binary mask with a probability $p$ dictating the presence of the original image, and $(\mathbb{1} - \mathbf{M}_p)$ represents the complementary mask applied to the Gaussian blurred image. $\mathbf{M}$ and $\mathbf{x}$ share the same dimensions. The parameter $p$ controls the contribution of the original image. A smaller $p$ results in more information being retained from the input image, leading to a more static output video. Conversely, as $p$ increases, the dynamic strength of the output is amplified, producing more motion.

Experimentally, we have found that this prior is critical to obtaining stable results, especially for relatively weak base T2I models such as Stable Diffusion V1.5~\cite{stablediffusion}. 

\section{Experiments}


\begin{table*}
\centering
\caption{Quantitative comparison and user study results on the MJHQ-30K Dataset.
In each column, the best number is highlighted in \textbf{bold}, while the second best one is \underline{underlined}.
}
\resizebox{1.\textwidth}{!}{
\renewcommand{\arraystretch}{1.3} 
\begin{tabular}{c|ccc|c|cccc|c}
\toprule
Method & \makecell[c]{Reusable to \\custom models} & \makecell[c]{Compatible with \\controllable tools} &               \makecell[c]{Multimodal \\inputs} & \makecell[c]{Trainable \\parameters} & DoverVQA $\uparrow$ &              CLIPTemp $\uparrow$ & FlowScore $\uparrow$ & WarpingError $\downarrow$ & User Study $\downarrow$\\

\midrule
Gen-2~\cite{gen2} & N/A & N/A & \Checkmark & N/A & 0.0800 & 0.9252 & 0.2490 & 12.5976 & 2.50\\
Pika~\cite{pika} & N/A & N/A & \Checkmark & N/A & 0.0691 & 0.9021 & 0.1316 & \textbf{9.4587} & 4.45\\
DynamiCrafter~\cite{xing2023dynamicrafter} & \XSolidBrush & \XSolidBrush & \XSolidBrush & 1.10B & 0.0848 & 0.9275 & 0.4185 & 14.3825 & 6.45\\
I2VGen-XL~\cite{2023i2vgenxl} & \XSolidBrush & \XSolidBrush & \Checkmark & 3.40B & 0.0459 & 0.6904 & 0.6111 & 12.6566 & 6.55\\
SVD~\cite{svd} & \XSolidBrush & \XSolidBrush & \XSolidBrush & 1.52B & 0.0847 & 0.9365 & 0.6883 & 20.8362 & 4.35\\
\textbf{I2V-Adapter (SD1.5)} & \Checkmark & \Checkmark & \Checkmark & {24M} & \underline{0.0873} & \underline{0.9379} & \underline{0.7069} & \underline{9.8967} & \underline{2.25}\\
\textbf{I2V-Adapter (SDXL)} & \Checkmark & \Checkmark & \Checkmark & {204M} & \textbf{0.0887} & \textbf{0.9389} & \textbf{0.7124} & 11.8388 & \textbf{1.45}\\
\bottomrule
\end{tabular}
\renewcommand{\arraystretch}{1} 
\label{tab:main}
}
\end{table*}

\subsection{Experimental Setup}
\paragraph{Datasets.}
We employ the WebVid-10M~\cite{bain2021webvid} dataset for training our models. This dataset encompasses roughly 10 million video clips in a resolution of 336 $\times$ 596, each averaging 18 seconds. Each video clip is coupled with a text description of the video content. A notable complication with WebVid-10M is the inclusion of watermarks, which inevitably appear in the generated videos. In order to generate videos without watermarks and at higher resolutions, we supplement our training data with additional self-collected video clips that are free of watermarks in a resolution of 720 $\times$ 1280. 
Finally, we employ the MJHQ-30K~\cite{mjhq30k} dataset as the benchmark for evaluation. This dataset encompasses 10 categories, each containing 3,000 samples, with each sample corresponding to an image and its text prompt.


\paragraph{Implementation details.}
We implement \sysname using two versions of AnimateDiff~\cite{guo2023animatediff} based on Stable Diffusion V1.5 (SD1.5)~\cite{stablediffusion} and SDXL~\cite{SDXL}, respectively. In both versions, we freeze the parameters of the pretrained models and only fine-tune the newly added modules. Specifically, for the SD1.5 version of AnimateDiff, we start by training on the WebVid-10M~\cite{bain2021webvid} dataset for one epoch, where input sequences of 16 frames are center-cropped to a resolution of 256 $\times$ 256. Subsequently, in the second phase, we train for another one epoch on a self-collected dataset, cropping in the same manner to a resolution of 512 $\times$ 512. During both phases, we use the AdamW optimizer~\cite{loshchilov2017adamw} with a fixed learning rate of 1e-4. Considering our training strategy does not predict for the first frame (the input image), we calculate the loss only for the subsequent 15 frames. Due to the necessity of higher resolution inputs for training the SDXL version, the implementation of this version is carried out exclusively on the self-collected dataset in the resolution of 512 $\times$ 512, with all other settings remaining unchanged.
Most results in our paper are based on the SDXL version.

\paragraph{Evaluation metrics}
We use four metrics to evaluate the results quantitatively, \ie, DoverVQA~\cite{wu2023dover}, CLIPTemp \cite{liu2023evalcrafter}, FlowScore~\cite{liu2023evalcrafter}, and WarpingError~\cite{liu2023evalcrafter}. 
DoverVQA indicates the video technical score, CLIPTemp reflects the consistency between the generated video and the reference image by calculating the average pairwise cosine similarity between input image and each frame. FlowScore measures the magnitude of motion within the video. WarpingError assesses the accuracy of the video's motion. FlowScore and WarpingError are obtained based on the optical flow computed via RAFT~\cite{RAFT}.
\revision{Additionally, we conduct a user study involving 10 participants. We evaluate all models using 20 test cases and ask participants to rank the results according to various factors,  including consistency with reference image, visual quality, and motion range. We report the results using Average User Ranking (AUR) as the user-preference metric.}
These metrics, reflecting model performance from various perspectives, enable a more comprehensive and impartial evaluation.

\subsection{Qualitative Results}

\begin{figure*}
    \centering
    \includegraphics[width=\linewidth]{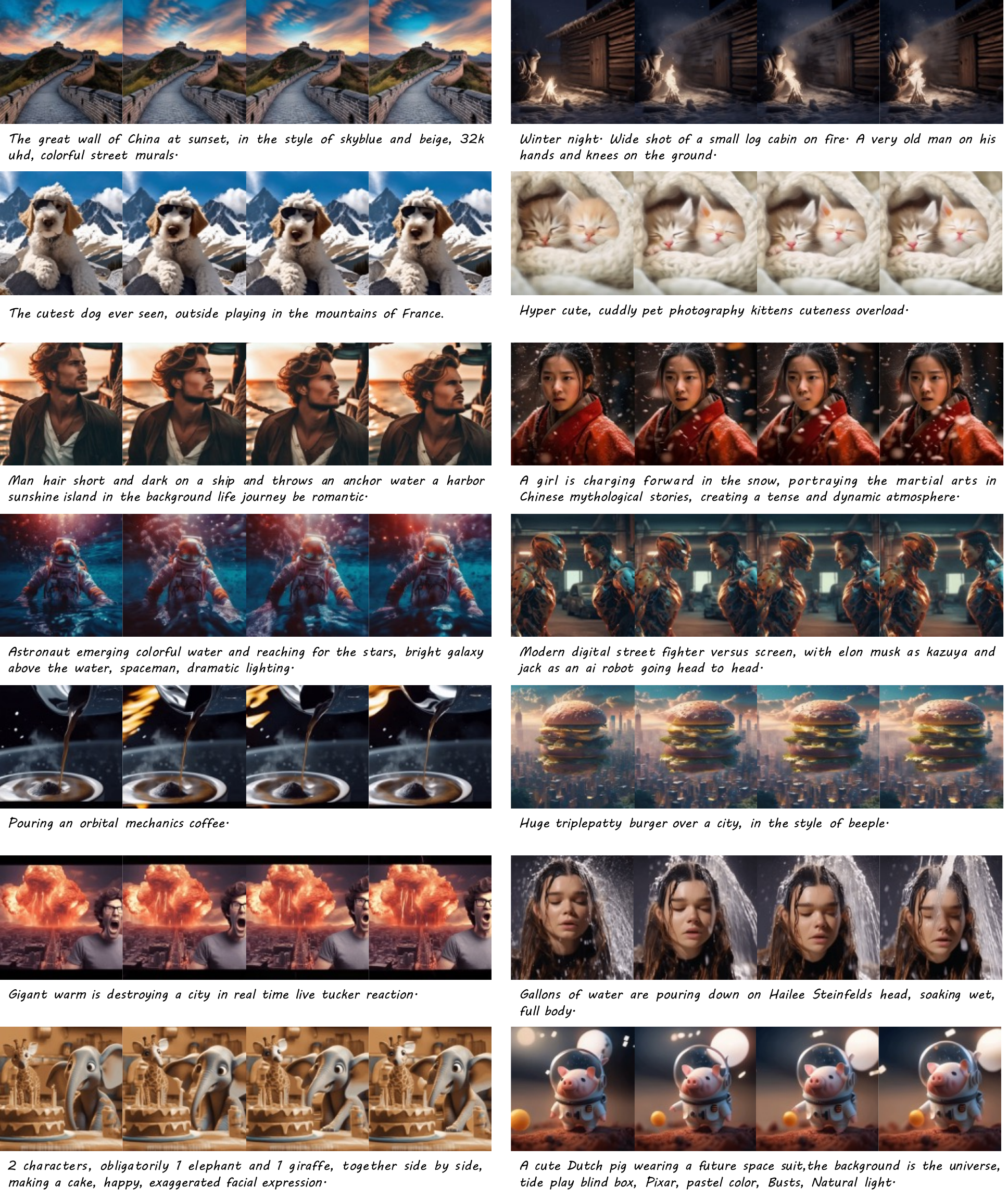}
    \caption{More videos of diverse categories generated by \sysname, with the text prompt at the bottom and the leftmost image as input.}
\label{fig:i2v_showcase}
\end{figure*}

\begin{figure*}
    \centering
    \includegraphics[width=\linewidth]{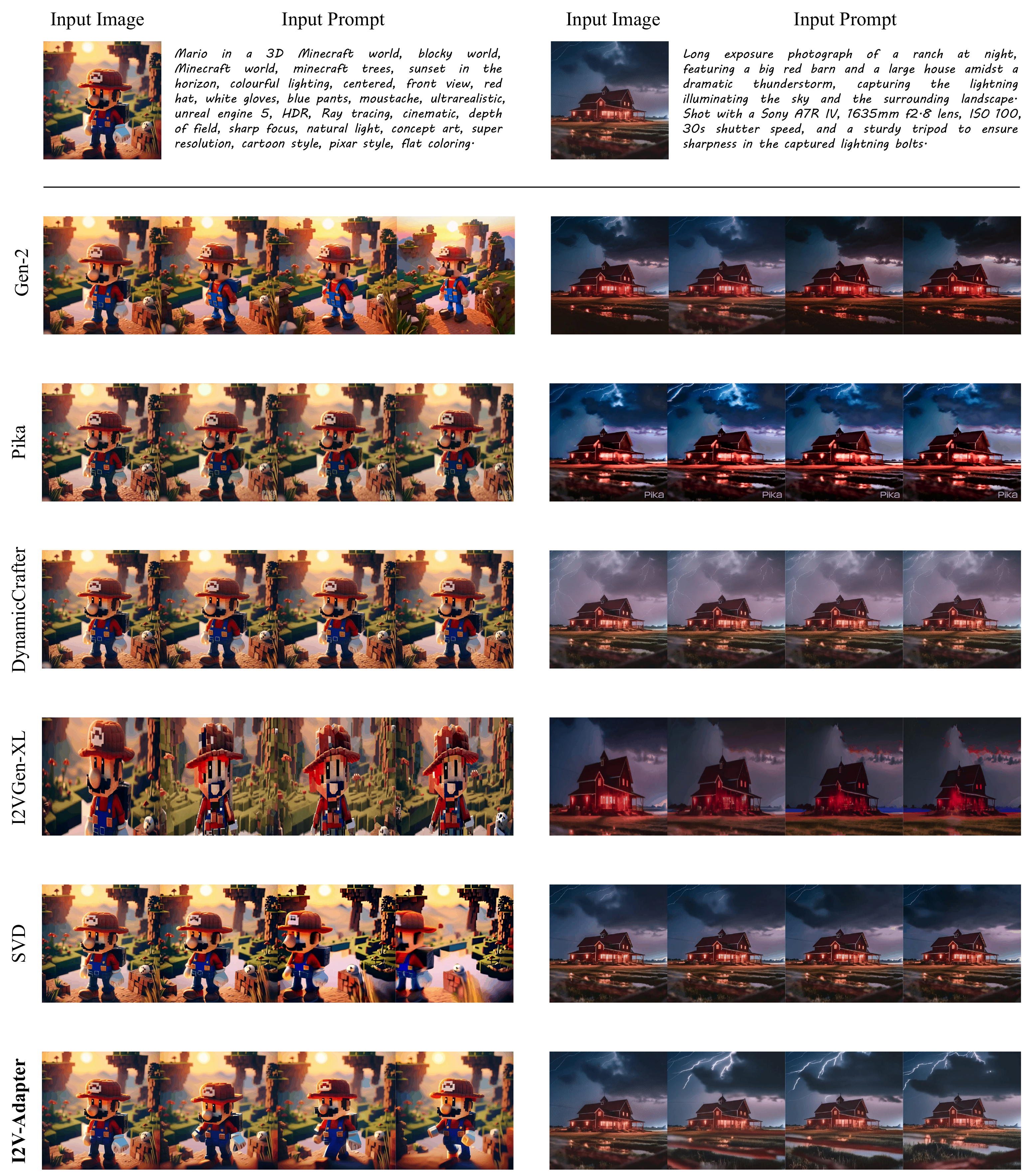}
    \caption{We compare our results with those obtained by SVD \cite{svd}, Gen-2 \cite{gen2}, Pika~\cite{pika}, and DynamiCrafter~\cite{xing2023dynamicrafter}, which are state-of-the-art I2V approaches. Results show that our model not only generates coherent and natural videos but also ensures consistency between the input image and subsequent frames. Moreover, our results demonstrate superior responsiveness to the input text prompt.}
\label{fig:i2v_cmp}
\end{figure*}

For qualitative evaluations, we conduct comparisons in an open-domain scenario with four existing I2V models, which encompass three community open-sourced models SVD~\cite{svd}, I2VGen-XL~\cite{2023i2vgenxl} and DynamicCrafter~\cite{xing2023dynamicrafter} and two commercial models Gen-2~\cite{gen2} and Pika~\cite{pika}. The results in Figure~\ref{fig:i2v_cmp} illustrate that our model generates coherent and natural videos while ensuring consistency between the input image and subsequent frames in terms of image identity. Moreover, our results demonstrate superior responsiveness to the input text prompt.
Figure~\ref{fig:i2v_showcase} presents additional results achieved by our method. \revision{Lastly, the results from the user study (the last column of Table~\ref{tab:main}) also demonstrate that our method surpasses the comparative solutions.}

\subsection{Quantitative Results}

\begin{figure}
    \centering
    \includegraphics[width=1.\linewidth]{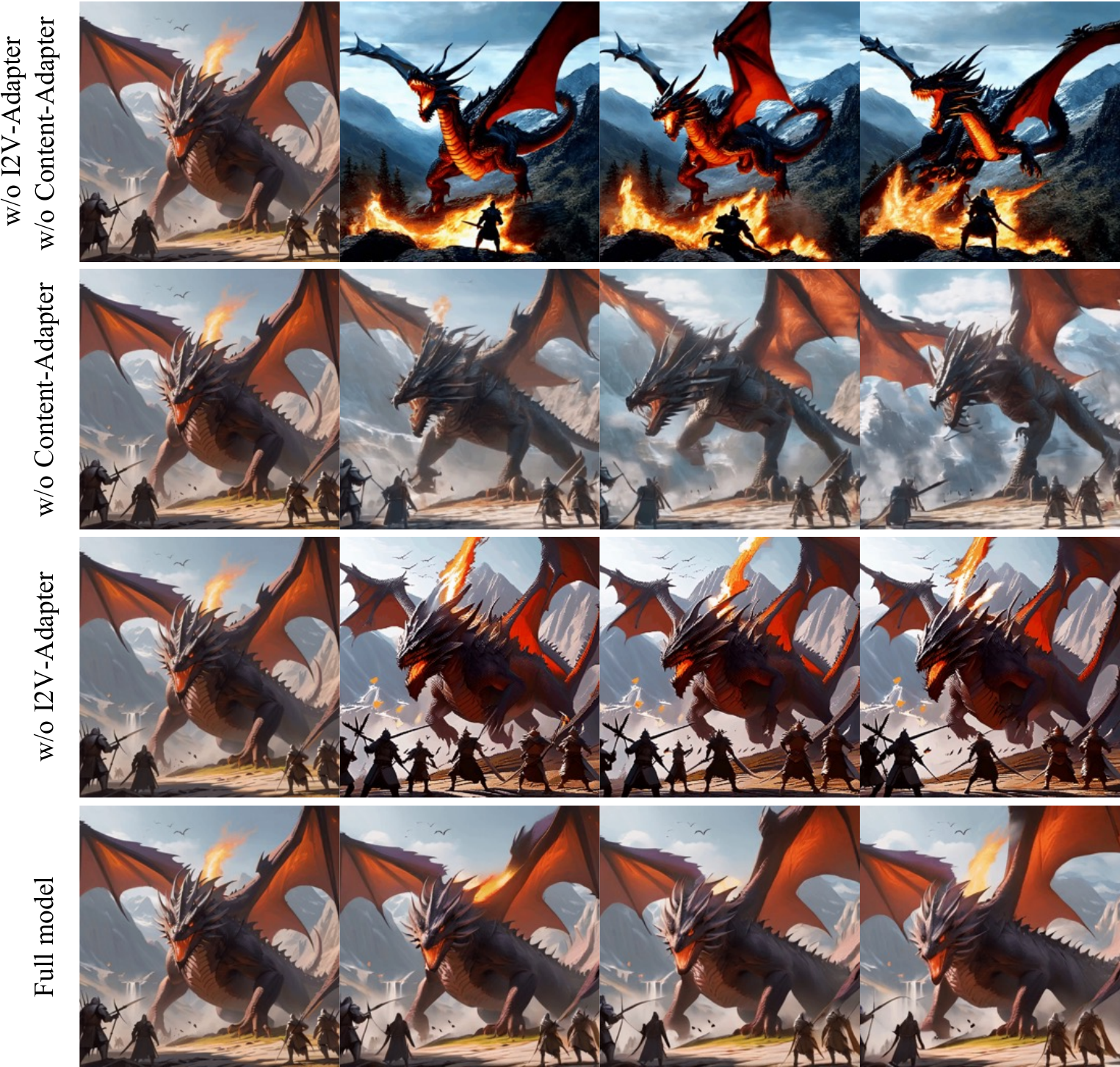}
    \caption{Qualitative ablation study results on model design.}
\label{fig:ablation_study}
\end{figure}

For quantitative analysis, we randomly sample 50 image-text pairs from MJHQ-30K dataset~\cite{mjhq30k} to conduct evaluations.
Table~\ref{tab:main} shows that our model achieves the highest aesthetic score, surpassing all competing models in terms of consistency with the reference image. Additionally, our generated videos exhibit the largest range of motion, and a relatively low warping error, indicating our model's ability to produce more dynamic videos while maintaining precise motion accuracy.

\subsection{Ablation Study}

\begin{figure}
    \centering
    \includegraphics[width=\linewidth]{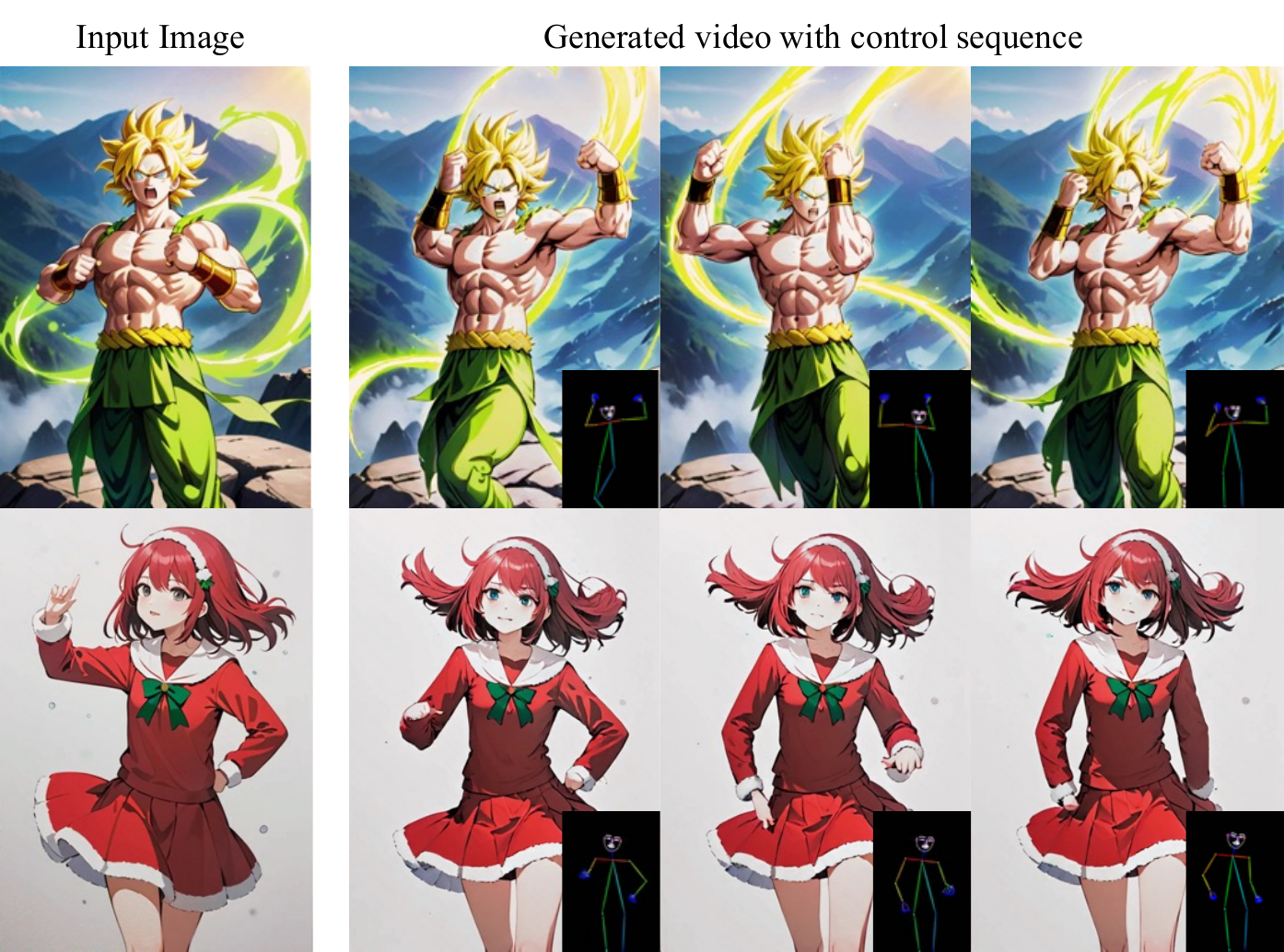}
    \caption{Our results with ControlNet~\cite{controlnet}. The pose control is shown as an inset at the bottom left for each frame.}
\label{fig:i2v_control}
\end{figure}





\paragraph{Model design}
To validate the effectiveness of the key components in our method, we use the proposed full model (the last row of Figure~\ref{fig:ablation_study}) as a baseline and design three alternative experimental configurations for comparison: (1) removing the I2V-Adapter module; (2) removing the Content-Adapter module; and (3) removing both of them. 
\revision{The corresponding qualitative and quantitative results are shown in Figure~\ref{fig:ablation_study} and Table~\ref{tab:model_design_ablation} respectively.}
The results reveal that the absence of the I2V-Adapter leads to significant loss of detail, as illustrated in the third row of Figure~\ref{fig:ablation_study}, where subsequent frames exhibit considerable changes in colors and finer details, underscoring the module's role in enhancing identity consistency. Conversely, in the absence of the Content-Adapter, the underlying information of subsequent frames remains largely consistent with the first frame, yet these frames tend to manifest instability, indicating a lack of global information guidance and control. Finally, the outcomes where both modules are omitted show poor performance, both in terms of detail preservation and global stability.

\begin{figure}
    \centering
    \includegraphics[width=1.0\linewidth]{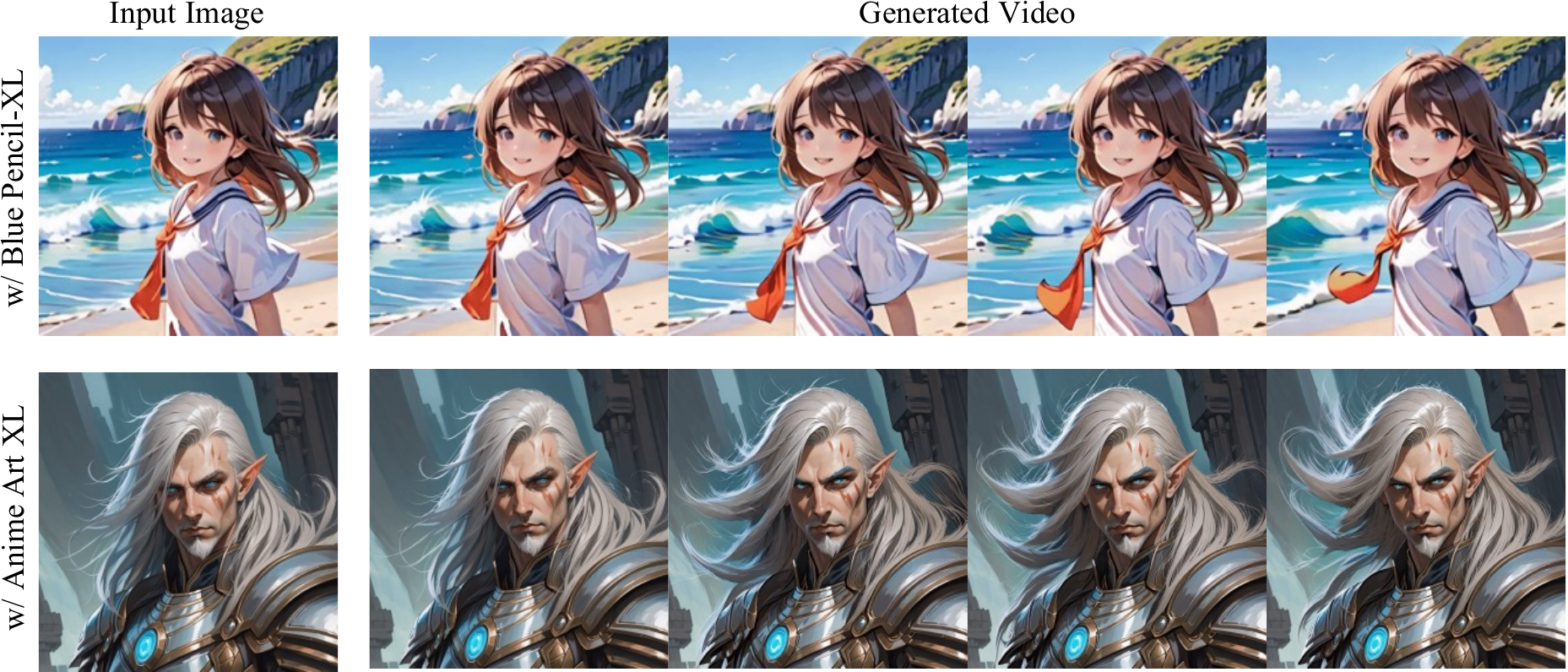}
    \caption{Our results with personalized T2I models~\cite{DreamBooth}.}
\label{fig:i2v_dreambooth}
\end{figure}

\paragraph{Frame Similarity Prior}
To verify the effectiveness of our Frame Similarity Prior (Section~\ref{sec:frame_similariy_prior}), we conduct experiments by adjusting the parameters $t_0$ and $p$ (see Table~\ref{tab:prior_ablation}). We observe that increasing the $p$ incrementally leads to a higher motion amplitude, as evidenced by the ascending values of the FlowScore metric. Simultaneously, the WarpingError metric exhibits a decreasing trend, indicating an enhancement in the stability of the generated video. Similarly, a gradual increase of the parameter $t_0$ also yields a comparable effect. These results indicate that our approach can effectively balance motion amplitude with the stability of the generated videos.


\begin{table}
\centering
\caption{Ablation study on parameters $t_0$ and $p$ of Frame Similarity Prior.}
\resizebox{0.48\textwidth}{!}{
\renewcommand{\arraystretch}{1.} 
\begin{tabular}{cc|ccccc}
\toprule
$t_0$ & $p$ & DoverVQA $\uparrow$ & CLIPTemp $\uparrow$ & FlowScore $\uparrow$ & WarpingError $\downarrow$ \\
\midrule
1.0 & 1.0 & 0.0900 & 0.9393 & 0.8395 & 11.7421 \\
0.9 & 1.0 & 0.0933 & 0.9445 & 0.4168 & 10.0771 \\
0.8 & 1.0 & 0.0940 & 0.9454 & 0.2712 & 9.1360 \\
0.7 & 1.0 & 0.0929 & 0.9485 & 0.1912 & 8.3508 \\
0.6 & 1.0 & 0.0925 & 0.9449 & 0.1428 & 7.6332 \\

\midrule
1.0 & 0.9 & 0.0904 & 0.9379 & 0.9895 & 12.0888 \\
1.0 & 0.7 & 0.0907 & 0.9389 & 0.9064 & 11.8403 \\
1.0 & 0.5 & 0.0894 & 0.9391 & 0.8046 & 11.6175 \\
1.0 & 0.3 & 0.0902 & 0.9404 & 0.7477 & 11.4993 \\
1.0 & 0.1 & 0.0901 & 0.9415 & 0.6918 & 11.3017 \\

\bottomrule
\end{tabular}
\renewcommand{\arraystretch}{1} 
\label{tab:prior_ablation}
}
\end{table}

\begin{table}
\centering
\caption{\revision{Quantitative results of ablation study on model design.}}
\resizebox{0.48\textwidth}{!}{
\renewcommand{\arraystretch}{1.} 
\begin{tabular}{c|ccccc}
\toprule
Model Design & DoverVQA $\uparrow$ & CLIPTemp $\uparrow$ & FlowScore $\uparrow$ & WarpingError $\downarrow$ \\
\midrule
w/o I2V-Adapter and Content-Adapter & 0.0264 & 0.7013 & 1.5671 & 24.5928 \\
w/o I2V-Adapter & 0.0428 & 0.8841 & 0.2684 & 13.7764 \\
w/o Content-Adapter & 0.0813 & 0.9158 & 1.3533 & 19.5557 \\
Full model & 0.0887 & 0.9389 & 0.7124 & 11.8388 \\
\bottomrule
\end{tabular}
\renewcommand{\arraystretch}{1} 
\label{tab:model_design_ablation}
}
\end{table}

\subsection{Additional Applications}

By keeping the parameters of the T2I models frozen, \sysname functions as a versatile plug-and-play module, allowing us to seamlessly integrate plugins from the open-source community. 
This section showcases two such applications.
First, given an input image and a sequence of poses, our model, in combination with ControlNet~\cite{controlnet}, generates a corresponding video (Figure~\ref{fig:i2v_control}).
Notably, our method adapts well to non-square resolutions different from the training resolution, underscoring its adaptability and generalization performance.
Second, with a reference image featuring a personalized style, our model can create a style-consistent video based on personalized T2I models (\eg, DreamBooth~\cite{DreamBooth}, LoRA~\cite{lora}), as shown in Figure~\ref{fig:i2v_dreambooth}.

\section{Conclusion and Future work}

In this paper, we introduce \sysname, a lightweight and efficient solution for text-guided image-to-video generation. Our method retains the prior knowledge of pretrained models. Our method propagates the unnoised input image to subsequent frames through a cross-frame attention mechanism, thereby ensuring the output subsequent frames are consistent with the input image. Furthermore, we introduce a novel Frame Similarity Prior to balance stability and motion magnitude of generated videos.

Our method has demonstrated effectiveness in open-domain image-to-video generation, producing high-quality videos with a reduced number of trainable parameters. In addition, our method retains the prior knowledge of pretrained models and the decoupled design ensures compatibility with existing community-driven personalized models and controlling tools, facilitating the creation of personalized and controllable videos without further training. 

\revision{Constrained by pretrained base models and available high-quality video data, our method can only generate 16-frame, 512x512 videos, which restricts the creation of longer or higher-resolution content.
}
In future work, we plan to incorporate frame interpolation and spatial super-resolution modules to generate videos of longer duration and higher resolution.

\begin{acks}
This work was supported by the National Natural Science Foundation of China (No. 62192785, No. 62372451, No. 62372082, No. U1936204, No. 62272125, No. 62306312, No. 62036011, No. 62192782 and No. 61721004, No. U2033210), the Project of Beijing Science and technology Committee (Project No. Z231100005923046), the Beijing Natural Science Foundation No. L223003, the Major Projects of Guangdong Education Department for Foundation Research and Applied Research (Grant: 2017KZDXM081, 2018KZDXM066), Guangdong Provincial University Innovation Team Project (Project No. 2020KCXTD045).
\end{acks}

\bibliographystyle{ACM-Reference-Format}
\bibliography{bib_ra}


\begin{thebibliography}{54}


\ifx \showCODEN    \undefined \def \showCODEN     #1{\unskip}     \fi
\ifx \showDOI      \undefined \def \showDOI       #1{#1}\fi
\ifx \showISBNx    \undefined \def \showISBNx     #1{\unskip}     \fi
\ifx \showISBNxiii \undefined \def \showISBNxiii  #1{\unskip}     \fi
\ifx \showISSN     \undefined \def \showISSN      #1{\unskip}     \fi
\ifx \showLCCN     \undefined \def \showLCCN      #1{\unskip}     \fi
\ifx \shownote     \undefined \def \shownote      #1{#1}          \fi
\ifx \showarticletitle \undefined \def \showarticletitle #1{#1}   \fi
\ifx \showURL      \undefined \def \showURL       {\relax}        \fi
\providecommand\bibfield[2]{#2}
\providecommand\bibinfo[2]{#2}
\providecommand\natexlab[1]{#1}
\providecommand\showeprint[2][]{arXiv:#2}

\bibitem[Bain et~al\mbox{.}(2021)]%
        {bain2021webvid}
\bibfield{author}{\bibinfo{person}{Max Bain}, \bibinfo{person}{Arsha Nagrani}, \bibinfo{person}{G{\"u}l Varol}, {and} \bibinfo{person}{Andrew Zisserman}.} \bibinfo{year}{2021}\natexlab{}.
\newblock \showarticletitle{Frozen in time: A joint video and image encoder for end-to-end retrieval}. In \bibinfo{booktitle}{\emph{Proceedings of the IEEE/CVF International Conference on Computer Vision}}. \bibinfo{pages}{1728--1738}.
\newblock


\bibitem[Blattmann et~al\mbox{.}(2023)]%
        {svd}
\bibfield{author}{\bibinfo{person}{Andreas Blattmann}, \bibinfo{person}{Tim Dockhorn}, \bibinfo{person}{Sumith Kulal}, \bibinfo{person}{Daniel Mendelevitch}, \bibinfo{person}{Maciej Kilian}, \bibinfo{person}{Dominik Lorenz}, \bibinfo{person}{Yam Levi}, \bibinfo{person}{Zion English}, \bibinfo{person}{Vikram Voleti}, \bibinfo{person}{Adam Letts}, {et~al\mbox{.}}} \bibinfo{year}{2023}\natexlab{}.
\newblock \showarticletitle{Stable video diffusion: Scaling latent video diffusion models to large datasets}.
\newblock \bibinfo{journal}{\emph{arXiv preprint arXiv:2311.15127}} (\bibinfo{year}{2023}).
\newblock


\bibitem[Cao et~al\mbox{.}(2023)]%
        {cao2023masactrl}
\bibfield{author}{\bibinfo{person}{Mingdeng Cao}, \bibinfo{person}{Xintao Wang}, \bibinfo{person}{Zhongang Qi}, \bibinfo{person}{Ying Shan}, \bibinfo{person}{Xiaohu Qie}, {and} \bibinfo{person}{Yinqiang Zheng}.} \bibinfo{year}{2023}\natexlab{}.
\newblock \showarticletitle{Masactrl: Tuning-free mutual self-attention control for consistent image synthesis and editing}. In \bibinfo{booktitle}{\emph{Proceedings of the IEEE/CVF International Conference on Computer Vision}}. \bibinfo{pages}{22560--22570}.
\newblock


\bibitem[Chen et~al\mbox{.}(2023c)]%
        {chen2023videocrafter1}
\bibfield{author}{\bibinfo{person}{Haoxin Chen}, \bibinfo{person}{Menghan Xia}, \bibinfo{person}{Yingqing He}, \bibinfo{person}{Yong Zhang}, \bibinfo{person}{Xiaodong Cun}, \bibinfo{person}{Shaoshu Yang}, \bibinfo{person}{Jinbo Xing}, \bibinfo{person}{Yaofang Liu}, \bibinfo{person}{Qifeng Chen}, \bibinfo{person}{Xintao Wang}, \bibinfo{person}{Chao Weng}, {and} \bibinfo{person}{Ying Shan}.} \bibinfo{year}{2023}\natexlab{c}.
\newblock \bibinfo{title}{VideoCrafter1: Open Diffusion Models for High-Quality Video Generation}.
\newblock
\newblock
\showeprint[arxiv]{2310.19512}~[cs.CV]


\bibitem[Chen et~al\mbox{.}(2023a)]%
        {chen2023livephoto}
\bibfield{author}{\bibinfo{person}{Xi Chen}, \bibinfo{person}{Zhiheng Liu}, \bibinfo{person}{Mengting Chen}, \bibinfo{person}{Yutong Feng}, \bibinfo{person}{Yu Liu}, \bibinfo{person}{Yujun Shen}, {and} \bibinfo{person}{Hengshuang Zhao}.} \bibinfo{year}{2023}\natexlab{a}.
\newblock \showarticletitle{LivePhoto: Real Image Animation with Text-guided Motion Control}.
\newblock \bibinfo{journal}{\emph{arXiv preprint arXiv:2312.02928}} (\bibinfo{year}{2023}).
\newblock


\bibitem[Chen et~al\mbox{.}(2023b)]%
        {chen2023seine}
\bibfield{author}{\bibinfo{person}{Xinyuan Chen}, \bibinfo{person}{Yaohui Wang}, \bibinfo{person}{Lingjun Zhang}, \bibinfo{person}{Shaobin Zhuang}, \bibinfo{person}{Xin Ma}, \bibinfo{person}{Jiashuo Yu}, \bibinfo{person}{Yali Wang}, \bibinfo{person}{Dahua Lin}, \bibinfo{person}{Yu Qiao}, {and} \bibinfo{person}{Ziwei Liu}.} \bibinfo{year}{2023}\natexlab{b}.
\newblock \showarticletitle{Seine: Short-to-long video diffusion model for generative transition and prediction}.
\newblock \bibinfo{journal}{\emph{arXiv preprint arXiv:2310.20700}} (\bibinfo{year}{2023}).
\newblock


\bibitem[Devlin et~al\mbox{.}(2019)]%
        {bert}
\bibfield{author}{\bibinfo{person}{Jacob Devlin}, \bibinfo{person}{Ming-Wei Chang}, \bibinfo{person}{Kenton Lee}, {and} \bibinfo{person}{Kristina Toutanova}.} \bibinfo{year}{2019}\natexlab{}.
\newblock \showarticletitle{{BERT}: Pre-training of Deep Bidirectional Transformers for Language Understanding}. In \bibinfo{booktitle}{\emph{Proceedings of the 2019 Conference of the North {A}merican Chapter of the Association for Computational Linguistics: Human Language Technologies, Volume 1 (Long and Short Papers)}}, \bibfield{editor}{\bibinfo{person}{Jill Burstein}, \bibinfo{person}{Christy Doran}, {and} \bibinfo{person}{Thamar Solorio}} (Eds.). \bibinfo{publisher}{Association for Computational Linguistics}, \bibinfo{address}{Minneapolis, Minnesota}, \bibinfo{pages}{4171--4186}.
\newblock
\urldef\tempurl%
\url{https://doi.org/10.18653/v1/N19-1423}
\showDOI{\tempurl}


\bibitem[Dhariwal and Nichol(2021)]%
        {dhariwal2021diffusion}
\bibfield{author}{\bibinfo{person}{Prafulla Dhariwal} {and} \bibinfo{person}{Alexander Nichol}.} \bibinfo{year}{2021}\natexlab{}.
\newblock \showarticletitle{Diffusion models beat gans on image synthesis}. In \bibinfo{booktitle}{\emph{Advances in neural information processing systems}}. \bibinfo{pages}{8780--8794}.
\newblock


\bibitem[Girdhar et~al\mbox{.}(2023)]%
        {emuvideo}
\bibfield{author}{\bibinfo{person}{Rohit Girdhar}, \bibinfo{person}{Mannat Singh}, \bibinfo{person}{Andrew Brown}, \bibinfo{person}{Quentin Duval}, \bibinfo{person}{Samaneh Azadi}, \bibinfo{person}{Sai~Saketh Rambhatla}, \bibinfo{person}{Akbar Shah}, \bibinfo{person}{Xi Yin}, \bibinfo{person}{Devi Parikh}, {and} \bibinfo{person}{Ishan Misra}.} \bibinfo{year}{2023}\natexlab{}.
\newblock \showarticletitle{Emu Video: Factorizing Text-to-Video Generation by Explicit Image Conditioning}.
\newblock \bibinfo{journal}{\emph{arXiv preprint arXiv:2311.10709}} (\bibinfo{year}{2023}).
\newblock


\bibitem[Guo et~al\mbox{.}(2023)]%
        {guo2023sparsectrl}
\bibfield{author}{\bibinfo{person}{Yuwei Guo}, \bibinfo{person}{Ceyuan Yang}, \bibinfo{person}{Anyi Rao}, \bibinfo{person}{Maneesh Agrawala}, \bibinfo{person}{Dahua Lin}, {and} \bibinfo{person}{Bo Dai}.} \bibinfo{year}{2023}\natexlab{}.
\newblock \showarticletitle{Sparsectrl: Adding sparse controls to text-to-video diffusion models}.
\newblock \bibinfo{journal}{\emph{arXiv preprint arXiv:2311.16933}} (\bibinfo{year}{2023}).
\newblock


\bibitem[Guo et~al\mbox{.}(2024)]%
        {guo2023animatediff}
\bibfield{author}{\bibinfo{person}{Yuwei Guo}, \bibinfo{person}{Ceyuan Yang}, \bibinfo{person}{Anyi Rao}, \bibinfo{person}{Zhengyang Liang}, \bibinfo{person}{Yaohui Wang}, \bibinfo{person}{Yu Qiao}, \bibinfo{person}{Maneesh Agrawala}, \bibinfo{person}{Dahua Lin}, {and} \bibinfo{person}{Bo Dai}.} \bibinfo{year}{2024}\natexlab{}.
\newblock \showarticletitle{AnimateDiff: Animate Your Personalized Text-to-Image Diffusion Models without Specific Tuning}. In \bibinfo{booktitle}{\emph{The Twelfth International Conference on Learning Representations}}.
\newblock


\bibitem[He et~al\mbox{.}(2016)]%
        {he2016resnet}
\bibfield{author}{\bibinfo{person}{Kaiming He}, \bibinfo{person}{Xiangyu Zhang}, \bibinfo{person}{Shaoqing Ren}, {and} \bibinfo{person}{Jian Sun}.} \bibinfo{year}{2016}\natexlab{}.
\newblock \showarticletitle{Deep residual learning for image recognition}. In \bibinfo{booktitle}{\emph{Proceedings of the IEEE conference on computer vision and pattern recognition}}. \bibinfo{pages}{770--778}.
\newblock


\bibitem[He et~al\mbox{.}(2022)]%
        {he2022lvdm}
\bibfield{author}{\bibinfo{person}{Yingqing He}, \bibinfo{person}{Tianyu Yang}, \bibinfo{person}{Yong Zhang}, \bibinfo{person}{Ying Shan}, {and} \bibinfo{person}{Qifeng Chen}.} \bibinfo{year}{2022}\natexlab{}.
\newblock \showarticletitle{Latent video diffusion models for high-fidelity video generation with arbitrary lengths}.
\newblock \bibinfo{journal}{\emph{arXiv preprint arXiv:2211.13221}} (\bibinfo{year}{2022}).
\newblock


\bibitem[Hertz et~al\mbox{.}(2022)]%
        {hertz2022p2p}
\bibfield{author}{\bibinfo{person}{Amir Hertz}, \bibinfo{person}{Ron Mokady}, \bibinfo{person}{Jay Tenenbaum}, \bibinfo{person}{Kfir Aberman}, \bibinfo{person}{Yael Pritch}, {and} \bibinfo{person}{Daniel Cohen-or}.} \bibinfo{year}{2022}\natexlab{}.
\newblock \showarticletitle{Prompt-to-Prompt Image Editing with Cross-Attention Control}. In \bibinfo{booktitle}{\emph{International Conference on Learning Representations (ICLR)}}.
\newblock


\bibitem[Ho et~al\mbox{.}(2022)]%
        {ho2022imagenvideo}
\bibfield{author}{\bibinfo{person}{Jonathan Ho}, \bibinfo{person}{William Chan}, \bibinfo{person}{Chitwan Saharia}, \bibinfo{person}{Jay Whang}, \bibinfo{person}{Ruiqi Gao}, \bibinfo{person}{Alexey Gritsenko}, \bibinfo{person}{Diederik~P Kingma}, \bibinfo{person}{Ben Poole}, \bibinfo{person}{Mohammad Norouzi}, \bibinfo{person}{David~J Fleet}, {et~al\mbox{.}}} \bibinfo{year}{2022}\natexlab{}.
\newblock \showarticletitle{Imagen video: High definition video generation with diffusion models}.
\newblock \bibinfo{journal}{\emph{arXiv preprint arXiv:2210.02303}} (\bibinfo{year}{2022}).
\newblock


\bibitem[Ho et~al\mbox{.}(2020)]%
        {ho2020denoising}
\bibfield{author}{\bibinfo{person}{Jonathan Ho}, \bibinfo{person}{Ajay Jain}, {and} \bibinfo{person}{Pieter Abbeel}.} \bibinfo{year}{2020}\natexlab{}.
\newblock \showarticletitle{Denoising diffusion probabilistic models}. In \bibinfo{booktitle}{\emph{Advances in neural information processing systems}}. \bibinfo{pages}{6840--6851}.
\newblock


\bibitem[Houlsby et~al\mbox{.}(2019)]%
        {houlsby2019adapter}
\bibfield{author}{\bibinfo{person}{Neil Houlsby}, \bibinfo{person}{Andrei Giurgiu}, \bibinfo{person}{Stanislaw Jastrzebski}, \bibinfo{person}{Bruna Morrone}, \bibinfo{person}{Quentin De~Laroussilhe}, \bibinfo{person}{Andrea Gesmundo}, \bibinfo{person}{Mona Attariyan}, {and} \bibinfo{person}{Sylvain Gelly}.} \bibinfo{year}{2019}\natexlab{}.
\newblock \showarticletitle{Parameter-efficient transfer learning for NLP}. In \bibinfo{booktitle}{\emph{International Conference on Machine Learning}}. PMLR, \bibinfo{pages}{2790--2799}.
\newblock


\bibitem[Hu et~al\mbox{.}(2022)]%
        {lora}
\bibfield{author}{\bibinfo{person}{Edward~J. Hu}, \bibinfo{person}{Yelong Shen}, \bibinfo{person}{Phillip Wallis}, \bibinfo{person}{Zeyuan Allen{-}Zhu}, \bibinfo{person}{Yuanzhi Li}, \bibinfo{person}{Shean Wang}, \bibinfo{person}{Lu Wang}, {and} \bibinfo{person}{Weizhu Chen}.} \bibinfo{year}{2022}\natexlab{}.
\newblock \showarticletitle{LoRA: Low-Rank Adaptation of Large Language Models}. In \bibinfo{booktitle}{\emph{International Conference on Learning Representations (ICLR)}}. \bibinfo{publisher}{OpenReview.net}.
\newblock


\bibitem[Hu et~al\mbox{.}(2023)]%
        {hu2023animate}
\bibfield{author}{\bibinfo{person}{Li Hu}, \bibinfo{person}{Xin Gao}, \bibinfo{person}{Peng Zhang}, \bibinfo{person}{Ke Sun}, \bibinfo{person}{Bang Zhang}, {and} \bibinfo{person}{Liefeng Bo}.} \bibinfo{year}{2023}\natexlab{}.
\newblock \showarticletitle{Animate anyone: Consistent and controllable image-to-video synthesis for character animation}.
\newblock \bibinfo{journal}{\emph{arXiv preprint arXiv:2311.17117}} (\bibinfo{year}{2023}).
\newblock


\bibitem[Kingma and Welling(2013)]%
        {kingma2013vae}
\bibfield{author}{\bibinfo{person}{Diederik~P Kingma} {and} \bibinfo{person}{Max Welling}.} \bibinfo{year}{2013}\natexlab{}.
\newblock \showarticletitle{Auto-encoding variational bayes}.
\newblock \bibinfo{journal}{\emph{arXiv preprint arXiv:1312.6114}} (\bibinfo{year}{2013}).
\newblock


\bibitem[Li et~al\mbox{.}(2023)]%
        {li2023videogen}
\bibfield{author}{\bibinfo{person}{Xin Li}, \bibinfo{person}{Wenqing Chu}, \bibinfo{person}{Ye Wu}, \bibinfo{person}{Weihang Yuan}, \bibinfo{person}{Fanglong Liu}, \bibinfo{person}{Qi Zhang}, \bibinfo{person}{Fu Li}, \bibinfo{person}{Haocheng Feng}, \bibinfo{person}{Errui Ding}, {and} \bibinfo{person}{Jingdong Wang}.} \bibinfo{year}{2023}\natexlab{}.
\newblock \showarticletitle{Videogen: A reference-guided latent diffusion approach for high definition text-to-video generation}.
\newblock \bibinfo{journal}{\emph{arXiv preprint arXiv:2309.00398}} (\bibinfo{year}{2023}).
\newblock


\bibitem[Liu et~al\mbox{.}(2023)]%
        {liu2023evalcrafter}
\bibfield{author}{\bibinfo{person}{Yaofang Liu}, \bibinfo{person}{Xiaodong Cun}, \bibinfo{person}{Xuebo Liu}, \bibinfo{person}{Xintao Wang}, \bibinfo{person}{Yong Zhang}, \bibinfo{person}{Haoxin Chen}, \bibinfo{person}{Yang Liu}, \bibinfo{person}{Tieyong Zeng}, \bibinfo{person}{Raymond Chan}, {and} \bibinfo{person}{Ying Shan}.} \bibinfo{year}{2023}\natexlab{}.
\newblock \showarticletitle{Evalcrafter: Benchmarking and evaluating large video generation models}.
\newblock \bibinfo{journal}{\emph{arXiv preprint arXiv:2310.11440}} (\bibinfo{year}{2023}).
\newblock


\bibitem[Loshchilov and Hutter(2017)]%
        {loshchilov2017adamw}
\bibfield{author}{\bibinfo{person}{Ilya Loshchilov} {and} \bibinfo{person}{Frank Hutter}.} \bibinfo{year}{2017}\natexlab{}.
\newblock \showarticletitle{Decoupled weight decay regularization}.
\newblock \bibinfo{journal}{\emph{arXiv preprint arXiv:1711.05101}} (\bibinfo{year}{2017}).
\newblock


\bibitem[Meng et~al\mbox{.}(2022)]%
        {sdedit}
\bibfield{author}{\bibinfo{person}{Chenlin Meng}, \bibinfo{person}{Yutong He}, \bibinfo{person}{Yang Song}, \bibinfo{person}{Jiaming Song}, \bibinfo{person}{Jiajun Wu}, \bibinfo{person}{Jun-Yan Zhu}, {and} \bibinfo{person}{Stefano Ermon}.} \bibinfo{year}{2022}\natexlab{}.
\newblock \showarticletitle{SDEdit: Guided Image Synthesis and Editing with Stochastic Differential Equations}. In \bibinfo{booktitle}{\emph{International Conference on Learning Representations (ICLR)}}.
\newblock


\bibitem[Mou et~al\mbox{.}(2023)]%
        {mou2023t2iadapter}
\bibfield{author}{\bibinfo{person}{Chong Mou}, \bibinfo{person}{Xintao Wang}, \bibinfo{person}{Liangbin Xie}, \bibinfo{person}{Yanze Wu}, \bibinfo{person}{Jian Zhang}, \bibinfo{person}{Zhongang Qi}, \bibinfo{person}{Ying Shan}, {and} \bibinfo{person}{Xiaohu Qie}.} \bibinfo{year}{2023}\natexlab{}.
\newblock \showarticletitle{T2i-adapter: Learning adapters to dig out more controllable ability for text-to-image diffusion models}.
\newblock \bibinfo{journal}{\emph{arXiv preprint arXiv:2302.08453}} (\bibinfo{year}{2023}).
\newblock


\bibitem[Mullan et~al\mbox{.}(2023)]%
        {Mullan_Hotshot-XL_2023}
\bibfield{author}{\bibinfo{person}{John Mullan}, \bibinfo{person}{Duncan Crawbuck}, {and} \bibinfo{person}{Aakash Sastry}.} \bibinfo{year}{2023}\natexlab{}.
\newblock \bibinfo{title}{{Hotshot-XL}}.
\newblock \bibinfo{howpublished}{\url{https://github.com/hotshotco/hotshot-xl}}.
\newblock


\bibitem[Oquab et~al\mbox{.}(2023)]%
        {oquab2023dinov2}
\bibfield{author}{\bibinfo{person}{Maxime Oquab}, \bibinfo{person}{Timoth{\'e}e Darcet}, \bibinfo{person}{Th{\'e}o Moutakanni}, \bibinfo{person}{Huy Vo}, \bibinfo{person}{Marc Szafraniec}, \bibinfo{person}{Vasil Khalidov}, \bibinfo{person}{Pierre Fernandez}, \bibinfo{person}{Daniel Haziza}, \bibinfo{person}{Francisco Massa}, \bibinfo{person}{Alaaeldin El-Nouby}, {et~al\mbox{.}}} \bibinfo{year}{2023}\natexlab{}.
\newblock \showarticletitle{Dinov2: Learning robust visual features without supervision}.
\newblock \bibinfo{journal}{\emph{arXiv preprint arXiv:2304.07193}} (\bibinfo{year}{2023}).
\newblock


\bibitem[{Pika}(2023)]%
        {pika}
\bibfield{author}{\bibinfo{person}{{Pika}}.} \bibinfo{year}{2023}\natexlab{}.
\newblock \bibinfo{title}{{Pika}}.
\newblock \bibinfo{howpublished}{\url{https://pika.art/}}.
\newblock


\bibitem[{PlaygroundAI}(2023)]%
        {mjhq30k}
\bibfield{author}{\bibinfo{person}{{PlaygroundAI}}.} \bibinfo{year}{2023}\natexlab{}.
\newblock \bibinfo{title}{{MJHQ-30K Dataset}}.
\newblock \bibinfo{howpublished}{\url{https://huggingface.co/datasets/playgroundai/MJHQ-30K}}.
\newblock


\bibitem[Podell et~al\mbox{.}(2023)]%
        {SDXL}
\bibfield{author}{\bibinfo{person}{Dustin Podell}, \bibinfo{person}{Zion English}, \bibinfo{person}{Kyle Lacey}, \bibinfo{person}{Andreas Blattmann}, \bibinfo{person}{Tim Dockhorn}, \bibinfo{person}{Jonas M{\"{u}}ller}, \bibinfo{person}{Joe Penna}, {and} \bibinfo{person}{Robin Rombach}.} \bibinfo{year}{2023}\natexlab{}.
\newblock \showarticletitle{{SDXL:} Improving Latent Diffusion Models for High-Resolution Image Synthesis}.
\newblock \bibinfo{journal}{\emph{CoRR}}  \bibinfo{volume}{abs/2307.01952} (\bibinfo{year}{2023}).
\newblock


\bibitem[Radford et~al\mbox{.}(2021)]%
        {clip}
\bibfield{author}{\bibinfo{person}{Alec Radford}, \bibinfo{person}{Jong~Wook Kim}, \bibinfo{person}{Chris Hallacy}, \bibinfo{person}{Aditya Ramesh}, \bibinfo{person}{Gabriel Goh}, \bibinfo{person}{Sandhini Agarwal}, \bibinfo{person}{Girish Sastry}, \bibinfo{person}{Amanda Askell}, \bibinfo{person}{Pamela Mishkin}, \bibinfo{person}{Jack Clark}, \bibinfo{person}{Gretchen Krueger}, {and} \bibinfo{person}{Ilya Sutskever}.} \bibinfo{year}{2021}\natexlab{}.
\newblock \showarticletitle{Learning Transferable Visual Models From Natural Language Supervision}. In \bibinfo{booktitle}{\emph{Proceedings of the 38th International Conference on Machine Learning, {ICML} 2021, 18-24 July 2021, Virtual Event}} \emph{(\bibinfo{series}{Proceedings of Machine Learning Research}, Vol.~\bibinfo{volume}{139})}, \bibfield{editor}{\bibinfo{person}{Marina Meila} {and} \bibinfo{person}{Tong Zhang}} (Eds.). \bibinfo{publisher}{{PMLR}}, \bibinfo{pages}{8748--8763}.
\newblock


\bibitem[Radford et~al\mbox{.}(2018)]%
        {radford2018gpt}
\bibfield{author}{\bibinfo{person}{Alec Radford}, \bibinfo{person}{Karthik Narasimhan}, \bibinfo{person}{Tim Salimans}, \bibinfo{person}{Ilya Sutskever}, {et~al\mbox{.}}} \bibinfo{year}{2018}\natexlab{}.
\newblock \bibinfo{booktitle}{\emph{Improving language understanding by generative pre-training}}.
\newblock \bibinfo{type}{{T}echnical {R}eport}. \bibinfo{institution}{OpenAI}.
\newblock


\bibitem[Rombach et~al\mbox{.}(2022)]%
        {stablediffusion}
\bibfield{author}{\bibinfo{person}{Robin Rombach}, \bibinfo{person}{Andreas Blattmann}, \bibinfo{person}{Dominik Lorenz}, \bibinfo{person}{Patrick Esser}, {and} \bibinfo{person}{Bj{\"{o}}rn Ommer}.} \bibinfo{year}{2022}\natexlab{}.
\newblock \showarticletitle{High-Resolution Image Synthesis with Latent Diffusion Models}. In \bibinfo{booktitle}{\emph{{IEEE/CVF} Conference on Computer Vision and Pattern Recognition, (CVPR)}}. \bibinfo{publisher}{{IEEE}}, \bibinfo{pages}{10674--10685}.
\newblock


\bibitem[Ronneberger et~al\mbox{.}(2015)]%
        {ronneberger2015unet}
\bibfield{author}{\bibinfo{person}{Olaf Ronneberger}, \bibinfo{person}{Philipp Fischer}, {and} \bibinfo{person}{Thomas Brox}.} \bibinfo{year}{2015}\natexlab{}.
\newblock \showarticletitle{U-net: Convolutional networks for biomedical image segmentation}. In \bibinfo{booktitle}{\emph{Medical Image Computing and Computer-Assisted Intervention--MICCAI 2015: 18th International Conference, Munich, Germany, October 5-9, 2015, Proceedings, Part III 18}}. Springer, \bibinfo{pages}{234--241}.
\newblock


\bibitem[Ruiz et~al\mbox{.}(2023)]%
        {DreamBooth}
\bibfield{author}{\bibinfo{person}{Nataniel Ruiz}, \bibinfo{person}{Yuanzhen Li}, \bibinfo{person}{Varun Jampani}, \bibinfo{person}{Yael Pritch}, \bibinfo{person}{Michael Rubinstein}, {and} \bibinfo{person}{Kfir Aberman}.} \bibinfo{year}{2023}\natexlab{}.
\newblock \showarticletitle{DreamBooth: Fine Tuning Text-to-Image Diffusion Models for Subject-Driven Generation}. In \bibinfo{booktitle}{\emph{{IEEE/CVF} Conference on Computer Vision and Pattern Recognition, {CVPR} 2023, Vancouver, BC, Canada, June 17-24, 2023}}. \bibinfo{publisher}{{IEEE}}, \bibinfo{pages}{22500--22510}.
\newblock


\bibitem[{RunwayAI}(2023)]%
        {gen2}
\bibfield{author}{\bibinfo{person}{{RunwayAI}}.} \bibinfo{year}{2023}\natexlab{}.
\newblock \bibinfo{title}{{Gen-2: The Next Step Forward for Generative AI}}.
\newblock \bibinfo{howpublished}{\url{https://research.runwayml.com/gen2}}.
\newblock


\bibitem[Song et~al\mbox{.}(2020a)]%
        {song2020denoising}
\bibfield{author}{\bibinfo{person}{Jiaming Song}, \bibinfo{person}{Chenlin Meng}, {and} \bibinfo{person}{Stefano Ermon}.} \bibinfo{year}{2020}\natexlab{a}.
\newblock \showarticletitle{Denoising Diffusion Implicit Models}. In \bibinfo{booktitle}{\emph{International Conference on Learning Representations (ICLR)}}.
\newblock


\bibitem[Song et~al\mbox{.}(2020b)]%
        {song2020score}
\bibfield{author}{\bibinfo{person}{Yang Song}, \bibinfo{person}{Jascha Sohl-Dickstein}, \bibinfo{person}{Diederik~P Kingma}, \bibinfo{person}{Abhishek Kumar}, \bibinfo{person}{Stefano Ermon}, {and} \bibinfo{person}{Ben Poole}.} \bibinfo{year}{2020}\natexlab{b}.
\newblock \showarticletitle{Score-Based Generative Modeling through Stochastic Differential Equations}. In \bibinfo{booktitle}{\emph{International Conference on Learning Representations (ICLR)}}.
\newblock


\bibitem[Teed and Deng(2020)]%
        {RAFT}
\bibfield{author}{\bibinfo{person}{Zachary Teed} {and} \bibinfo{person}{Jia Deng}.} \bibinfo{year}{2020}\natexlab{}.
\newblock \showarticletitle{{RAFT:} Recurrent All-Pairs Field Transforms for Optical Flow}. In \bibinfo{booktitle}{\emph{{ECCV} {(2)}}} \emph{(\bibinfo{series}{Lecture Notes in Computer Science}, Vol.~\bibinfo{volume}{12347})}. \bibinfo{publisher}{Springer}, \bibinfo{pages}{402--419}.
\newblock


\bibitem[Tumanyan et~al\mbox{.}(2023)]%
        {tumanyan2023plug}
\bibfield{author}{\bibinfo{person}{Narek Tumanyan}, \bibinfo{person}{Michal Geyer}, \bibinfo{person}{Shai Bagon}, {and} \bibinfo{person}{Tali Dekel}.} \bibinfo{year}{2023}\natexlab{}.
\newblock \showarticletitle{Plug-and-play diffusion features for text-driven image-to-image translation}. In \bibinfo{booktitle}{\emph{Proceedings of the IEEE/CVF Conference on Computer Vision and Pattern Recognition}}. \bibinfo{pages}{1921--1930}.
\newblock


\bibitem[Vaswani et~al\mbox{.}(2017)]%
        {vaswani2017attention}
\bibfield{author}{\bibinfo{person}{Ashish Vaswani}, \bibinfo{person}{Noam Shazeer}, \bibinfo{person}{Niki Parmar}, \bibinfo{person}{Jakob Uszkoreit}, \bibinfo{person}{Llion Jones}, \bibinfo{person}{Aidan~N Gomez}, \bibinfo{person}{{\L}ukasz Kaiser}, {and} \bibinfo{person}{Illia Polosukhin}.} \bibinfo{year}{2017}\natexlab{}.
\newblock \showarticletitle{Attention is all you need}. In \bibinfo{booktitle}{\emph{Advances in neural information processing systems}}.
\newblock


\bibitem[Voleti et~al\mbox{.}(2022)]%
        {voleti2022mcvd}
\bibfield{author}{\bibinfo{person}{Vikram Voleti}, \bibinfo{person}{Alexia Jolicoeur-Martineau}, {and} \bibinfo{person}{Chris Pal}.} \bibinfo{year}{2022}\natexlab{}.
\newblock \showarticletitle{MCVD-masked conditional video diffusion for prediction, generation, and interpolation}. In \bibinfo{booktitle}{\emph{Advances in Neural Information Processing Systems}}. \bibinfo{pages}{23371--23385}.
\newblock


\bibitem[Wang et~al\mbox{.}(2023b)]%
        {wang2023modelscope}
\bibfield{author}{\bibinfo{person}{Jiuniu Wang}, \bibinfo{person}{Hangjie Yuan}, \bibinfo{person}{Dayou Chen}, \bibinfo{person}{Yingya Zhang}, \bibinfo{person}{Xiang Wang}, {and} \bibinfo{person}{Shiwei Zhang}.} \bibinfo{year}{2023}\natexlab{b}.
\newblock \showarticletitle{Modelscope text-to-video technical report}.
\newblock \bibinfo{journal}{\emph{arXiv preprint arXiv:2308.06571}} (\bibinfo{year}{2023}).
\newblock


\bibitem[Wang et~al\mbox{.}(2023c)]%
        {wang2023videocomposer}
\bibfield{author}{\bibinfo{person}{Xiang Wang}, \bibinfo{person}{Hangjie Yuan}, \bibinfo{person}{Shiwei Zhang}, \bibinfo{person}{Dayou Chen}, \bibinfo{person}{Jiuniu Wang}, \bibinfo{person}{Yingya Zhang}, \bibinfo{person}{Yujun Shen}, \bibinfo{person}{Deli Zhao}, {and} \bibinfo{person}{Jingren Zhou}.} \bibinfo{year}{2023}\natexlab{c}.
\newblock \showarticletitle{VideoComposer: Compositional Video Synthesis with Motion Controllability}. In \bibinfo{booktitle}{\emph{Advances in Neural Information Processing Systems}}. \bibinfo{pages}{7594--7611}.
\newblock


\bibitem[Wang et~al\mbox{.}(2023a)]%
        {wang2023lavie}
\bibfield{author}{\bibinfo{person}{Yaohui Wang}, \bibinfo{person}{Xinyuan Chen}, \bibinfo{person}{Xin Ma}, \bibinfo{person}{Shangchen Zhou}, \bibinfo{person}{Ziqi Huang}, \bibinfo{person}{Yi Wang}, \bibinfo{person}{Ceyuan Yang}, \bibinfo{person}{Yinan He}, \bibinfo{person}{Jiashuo Yu}, \bibinfo{person}{Peiqing Yang}, {et~al\mbox{.}}} \bibinfo{year}{2023}\natexlab{a}.
\newblock \showarticletitle{Lavie: High-quality video generation with cascaded latent diffusion models}.
\newblock \bibinfo{journal}{\emph{arXiv preprint arXiv:2309.15103}} (\bibinfo{year}{2023}).
\newblock


\bibitem[Witteveen and Andrews(2022)]%
        {witteveen2022investigating}
\bibfield{author}{\bibinfo{person}{Sam Witteveen} {and} \bibinfo{person}{Martin Andrews}.} \bibinfo{year}{2022}\natexlab{}.
\newblock \showarticletitle{Investigating prompt engineering in diffusion models}.
\newblock \bibinfo{journal}{\emph{arXiv preprint arXiv:2211.15462}} (\bibinfo{year}{2022}).
\newblock


\bibitem[Wu et~al\mbox{.}(2023b)]%
        {wu2023dover}
\bibfield{author}{\bibinfo{person}{Haoning Wu}, \bibinfo{person}{Erli Zhang}, \bibinfo{person}{Liang Liao}, \bibinfo{person}{Chaofeng Chen}, \bibinfo{person}{Jingwen~Hou Hou}, \bibinfo{person}{Annan Wang}, \bibinfo{person}{Wenxiu~Sun Sun}, \bibinfo{person}{Qiong Yan}, {and} \bibinfo{person}{Weisi Lin}.} \bibinfo{year}{2023}\natexlab{b}.
\newblock \showarticletitle{Exploring Video Quality Assessment on User Generated Contents from Aesthetic and Technical Perspectives}. In \bibinfo{booktitle}{\emph{International Conference on Computer Vision (ICCV)}}.
\newblock


\bibitem[Wu et~al\mbox{.}(2023a)]%
        {wu2023lamp}
\bibfield{author}{\bibinfo{person}{Ruiqi Wu}, \bibinfo{person}{Liangyu Chen}, \bibinfo{person}{Tong Yang}, \bibinfo{person}{Chunle Guo}, \bibinfo{person}{Chongyi Li}, {and} \bibinfo{person}{Xiangyu Zhang}.} \bibinfo{year}{2023}\natexlab{a}.
\newblock \showarticletitle{Lamp: Learn a motion pattern for few-shot-based video generation}.
\newblock \bibinfo{journal}{\emph{arXiv preprint arXiv:2310.10769}} (\bibinfo{year}{2023}).
\newblock


\bibitem[Xing et~al\mbox{.}(2023)]%
        {xing2023dynamicrafter}
\bibfield{author}{\bibinfo{person}{Jinbo Xing}, \bibinfo{person}{Menghan Xia}, \bibinfo{person}{Yong Zhang}, \bibinfo{person}{Haoxin Chen}, \bibinfo{person}{Xintao Wang}, \bibinfo{person}{Tien-Tsin Wong}, {and} \bibinfo{person}{Ying Shan}.} \bibinfo{year}{2023}\natexlab{}.
\newblock \showarticletitle{Dynamicrafter: Animating open-domain images with video diffusion priors}.
\newblock \bibinfo{journal}{\emph{arXiv preprint arXiv:2310.12190}} (\bibinfo{year}{2023}).
\newblock


\bibitem[Xu et~al\mbox{.}(2023)]%
        {xu2023magicanimate}
\bibfield{author}{\bibinfo{person}{Zhongcong Xu}, \bibinfo{person}{Jianfeng Zhang}, \bibinfo{person}{Jun~Hao Liew}, \bibinfo{person}{Hanshu Yan}, \bibinfo{person}{Jia-Wei Liu}, \bibinfo{person}{Chenxu Zhang}, \bibinfo{person}{Jiashi Feng}, {and} \bibinfo{person}{Mike~Zheng Shou}.} \bibinfo{year}{2023}\natexlab{}.
\newblock \showarticletitle{Magicanimate: Temporally consistent human image animation using diffusion model}.
\newblock \bibinfo{journal}{\emph{arXiv preprint arXiv:2311.16498}} (\bibinfo{year}{2023}).
\newblock


\bibitem[Ye et~al\mbox{.}(2023)]%
        {ye2023ipadapter}
\bibfield{author}{\bibinfo{person}{Hu Ye}, \bibinfo{person}{Jun Zhang}, \bibinfo{person}{Sibo Liu}, \bibinfo{person}{Xiao Han}, {and} \bibinfo{person}{Wei Yang}.} \bibinfo{year}{2023}\natexlab{}.
\newblock \showarticletitle{Ip-adapter: Text compatible image prompt adapter for text-to-image diffusion models}.
\newblock \bibinfo{journal}{\emph{arXiv preprint arXiv:2308.06721}} (\bibinfo{year}{2023}).
\newblock


\bibitem[Zeng et~al\mbox{.}(2023)]%
        {zeng2023makepixeldance}
\bibfield{author}{\bibinfo{person}{Yan Zeng}, \bibinfo{person}{Guoqiang Wei}, \bibinfo{person}{Jiani Zheng}, \bibinfo{person}{Jiaxin Zou}, \bibinfo{person}{Yang Wei}, \bibinfo{person}{Yuchen Zhang}, {and} \bibinfo{person}{Hang Li}.} \bibinfo{year}{2023}\natexlab{}.
\newblock \showarticletitle{Make Pixels Dance: High-Dynamic Video Generation}.
\newblock \bibinfo{journal}{\emph{arXiv preprint arXiv:2311.10982}} (\bibinfo{year}{2023}).
\newblock


\bibitem[Zhang et~al\mbox{.}(2023a)]%
        {controlnet}
\bibfield{author}{\bibinfo{person}{Lvmin Zhang}, \bibinfo{person}{Anyi Rao}, {and} \bibinfo{person}{Maneesh Agrawala}.} \bibinfo{year}{2023}\natexlab{a}.
\newblock \showarticletitle{Adding conditional control to text-to-image diffusion models}. In \bibinfo{booktitle}{\emph{Proceedings of the IEEE/CVF International Conference on Computer Vision}}. \bibinfo{pages}{3836--3847}.
\newblock


\bibitem[Zhang et~al\mbox{.}(2023b)]%
        {2023i2vgenxl}
\bibfield{author}{\bibinfo{person}{Shiwei Zhang}, \bibinfo{person}{Jiayu Wang}, \bibinfo{person}{Yingya Zhang}, \bibinfo{person}{Kang Zhao}, \bibinfo{person}{Hangjie Yuan}, \bibinfo{person}{Zhiwu Qing}, \bibinfo{person}{Xiang Wang}, \bibinfo{person}{Deli Zhao}, {and} \bibinfo{person}{Jingren Zhou}.} \bibinfo{year}{2023}\natexlab{b}.
\newblock \showarticletitle{I2VGen-XL: High-Quality Image-to-Video Synthesis via Cascaded Diffusion Models}.
\newblock \bibinfo{journal}{\emph{arXiv preprint arXiv:2311.04145}} (\bibinfo{year}{2023}).
\newblock


\end{thebibliography}

\newpage

\end{document}


\appendix
\title{Towards Practical Capture of High-Fidelity Relightable Avatars -- Supplementary Materials}

\renewcommand{\onesixthfigurewidth}{0.163}

\begin{teaserfigure}
    \centering
    \begin{minipage}[t]{\linewidth}
        \centering
        \includegraphics[width=\onesixthfigurewidth\linewidth]{images/ablation/light1_resize.png}
        \includegraphics[width=\onesixthfigurewidth\linewidth]{images/ablation/f3_nl_crop.png}
        \includegraphics[width=\onesixthfigurewidth\linewidth]{images/ablation/f3_env_crop.png}
        \includegraphics[width=\onesixthfigurewidth\linewidth]{images/ablation/f3_lcl_crop.png}
        \includegraphics[width=\onesixthfigurewidth\linewidth]{images/ablation/f3_ts_crop.png}
        \includegraphics[width=\onesixthfigurewidth\linewidth]{images/ablation/f3_ours_crop.png}
        \includegraphics[width=\onesixthfigurewidth\linewidth]{images/ablation/light2_resize.png}
        \includegraphics[width=\onesixthfigurewidth\linewidth]{images/ablation/f1_nl_crop.png}
        \includegraphics[width=\onesixthfigurewidth\linewidth]{images/ablation/f1_env_crop.png}
        \includegraphics[width=\onesixthfigurewidth\linewidth]{images/ablation/f1_lcl_crop.png}
        \includegraphics[width=\onesixthfigurewidth\linewidth]{images/ablation/f1_ts_crop.png}
        \includegraphics[width=\onesixthfigurewidth\linewidth]{images/ablation/f1_ours_crop.png}
        \begin{minipage}[t]{\onesixthfigurewidth\linewidth}
            \centering
            \subfloat{Light}
        \end{minipage}
        \begin{minipage}[t]{\onesixthfigurewidth\linewidth}
            \centering
            \subfloat{NL}
        \end{minipage}
        \begin{minipage}[t]{\onesixthfigurewidth\linewidth}
            \centering
            \subfloat{NL + ENV}
        \end{minipage}
        \begin{minipage}[t]{\onesixthfigurewidth\linewidth}
            \centering
            \subfloat{NL + LCL}
        \end{minipage}
        \begin{minipage}[t]{\onesixthfigurewidth\linewidth}
            \centering
            \subfloat{NL + TS}
        \end{minipage}
        \begin{minipage}[t]{\onesixthfigurewidth\linewidth}
            \centering
            \subfloat{Ours}
        \end{minipage}
    \end{minipage}
    \captionof{figure}{Ablation study results about our physically inspired linear light branch for the appearance decoder $\colordecoder$.
     The input environment maps are shown on the left.
    The relighting results of four alternative baseline approaches (see detailed explanations in Section 5.3 of the main text) and ours are shown on the right.
    }
\label{fig:ablation}
\end{teaserfigure}

\renewcommand{\shortauthors}{H. Yang, M. Zheng, W. Feng, H. Huang, Y.-K. Lai, P. Wan, Z. Wang, and C. Ma}

\maketitle

\revision{
\section{Additional Results}

Figure~\ref{fig:ablation} shows additional ablation study results using two environment maps, which demonstrate that the linear lighting branch of our appearance decoder $\colordecoder$ can significantly enhance the generalization performance for relighting.
}

\begin{figure*}
\centering
    \includegraphics[width=\linewidth]{figure/networks.pdf}
    \caption{Detailed architecture of our motion encoder $\motionencoder$, opacity decoder $\opacitydecoder$, and appearance decoder $\colordecoder$. The convolutional layer is represented as $Conv(inchs, outchs, kernel size, stride)$, where $inchs$ is the number of input channels and $outchs$ is the number of output channels. The representation of transposed convolutional layers are similar. The fully connected layer is represented as $FC(inchs, outchs)$. 
    The architecture of the transformation decoder $\transformdecoder$ is similar to the opacity decoder, except that there is no activation layer at the end.
    The mesh decoder $\meshdecoder$ is a three-layer MLP with LeakyReLU activation layers, which is omitted in this figure.}
\label{fig:networks}
\end{figure*}

\section{Implementation Details}

\paragraph{Network architectures and hyperparameters.}
We provide detailed architectures of our neural networks in Figure~\ref{fig:networks}.
The values of hyperparameters in our implementation are provided in Table~\ref{tab:hyperparameter}, which are identical in all our experiments.

\begin{table}[]
\centering
\caption{Values of our hyperparameters.}
\begin{tabular}{cr|cr|cr}
\hline
Parameter & Value & Parameter & Value & Parameter & Value \\ \hline
$N_{mesh}$     & 7306  & $N_{prim}$     & 16384 & $M$         & 8     \\
${\lambda}_\mathrm{VGG}$    & 0.1   & ${\lambda}_\mathrm{GAN}$    & 0.005 & ${\lambda}_\mathrm{Lap}$    & 0.01  \\
${\lambda}_{pR}$     & 10    & ${\lambda}_{vol}$    & 0.01  & ${\lambda}_\mathrm{KLD}$    & 0.001 \\ \hline
\end{tabular}
\label{tab:hyperparameter}%
\end{table}

\begin{figure}
    \centering
    \begin{minipage}[t]{0.99\linewidth}
        \centering
        \includegraphics[width=\linewidth]{figure/lightposition.png}
    \end{minipage}
    \caption{The projected light positions on an environment map. Green circles: positions of 356 lighting units projected on an environment map in the longitude-latitude format. Blue lines: edges of the Voronoi diagram.}
\label{fig:lightposition}
\end{figure}

\paragraph{Environment map relighting.}
Given a high dynamic range environment map in the longitude-latitude format, we extract the lighting condition $\light$ for our appearance decoder. Specifically, we project the position of each light unit of the Light Stage onto the environment map image and split the space using a Voronoi diagram~\cite{aurenhammer1991voronoi}. The corresponding value of $\light$ is set according to the weighted-average pixel values in the cell. We show the projected light positions in Figure~\ref{fig:lightposition}.

\bibliographystyle{ACM-Reference-Format}
\bibliography{bib_ra}